\documentclass[letterpaper]{article} 
\usepackage{aaai2026}  
\usepackage{times}  
\usepackage{helvet}  
\usepackage{courier}  
\usepackage[hyphens]{url}  
\usepackage{graphicx} 
\usepackage{natbib}  
\usepackage{caption} 
\usepackage{algorithm}
\usepackage{algorithmic}

\newcommand{\etal}{\textit{et al.}}
\usepackage{multirow}
\usepackage{booktabs}
\usepackage{picins}
\usepackage{lineno}
\usepackage{colortbl}
\usepackage{amsmath,amssymb,amsfonts}
\usepackage{makecell}

\usepackage{array}
\usepackage[caption=false,font=normalsize,labelfont=sf,textfont=sf]{subfig}
\usepackage{textcomp}
\usepackage{verbatim}

\usepackage{cite}
\usepackage{newfloat}
\usepackage{listings}
\DeclareCaptionStyle{ruled}{labelfont=normalfont,labelsep=colon,strut=off} 
\floatstyle{ruled}
\newfloat{listing}{tb}{lst}{}
\floatname{listing}{Listing}
\title{Visual-Friendly Concept Protection via Selective Adversarial Perturbations}

\author{
    Xiaoyue Mi\textsuperscript{\rm 1,2},
    Fan Tang\textsuperscript{\rm 1,2}\thanks{Corresponding author},
    You Wu\textsuperscript{\rm 1,2},
    Juan Cao\textsuperscript{\rm 1,2},
    Peng Li\textsuperscript{\rm 3},
    Yang Liu\textsuperscript{\rm 3,4}
}
\affiliations{
    \textsuperscript{\rm 1}Institute of Computing Technology, Chinese Academy of Sciences\\
    \textsuperscript{\rm 2}University of Chinese Academy of Sciences\\
    \textsuperscript{\rm 3}Institute for AI Industry Research (AIR), Tsinghua University\\
    \textsuperscript{\rm 4}Department of Computer Science \& Technology, Tsinghua University\\
    mixiaoyue19s@ict.ac.cn, tfan.108@gmail.com, wwwuyou99@gmail.com, caojuan@ict.ac.cn, pengli09@gmail.com, liuyang2011@tsinghua.edu.cn
}

\begin{document}

\maketitle

\begin{abstract}
Personalized concept generation by tuning diffusion models with a few images raises potential legal and ethical concerns regarding privacy and intellectual property rights.
Researchers attempt to prevent malicious personalization using adversarial perturbations.
However, previous efforts have mainly focused on the effectiveness of protection while neglecting the visibility of perturbations.
They utilize global adversarial perturbations, which introduce noticeable alterations to original images and significantly degrade visual quality.
In this work, we propose the \underline{V}isual-Friendly \underline{C}oncept \underline{Pro}tection (VCPro) framework, which prioritizes the protection of key concepts chosen by the image owner through adversarial perturbations with lower perceptibility.
To ensure these perturbations are as inconspicuous as possible, we introduce a relaxed optimization objective to identify the least perceptible yet effective adversarial perturbations, solved using the Lagrangian multiplier method.
Qualitative and quantitative experiments validate that VCPro achieves a better trade-off between the visibility of perturbations and protection effectiveness, effectively prioritizing the protection of target concepts in images with less perceptible perturbations.
\end{abstract}


\section{Introduction}
With the advent of popular image generative models~\cite{ho2020denoising,rombach2022high,song2020denoising} such as Stable Diffusion~\cite{rombach2022high} and GPT-4o~\cite{achiam2023gpt}, people lacking expertise in drawing or photography can effortlessly create realistic or artistic works using simple textual descriptions.
However, the success of these models has raised significant concerns about privacy, intellectual property rights, and various legal and ethical issues~\cite{luo2024exploring}.
For example, an adversary could easily generate sensitive specific concepts, such as personal fake images, or imitate renowned artworks for commercial purposes, using only a few reference images and concept personalization techniques such as Textual Inversion~\cite{gal2022image} or DreamBooth~\cite{ruiz2023dreambooth}.
\begin{figure*}
    \centering
       \includegraphics[width=\linewidth]{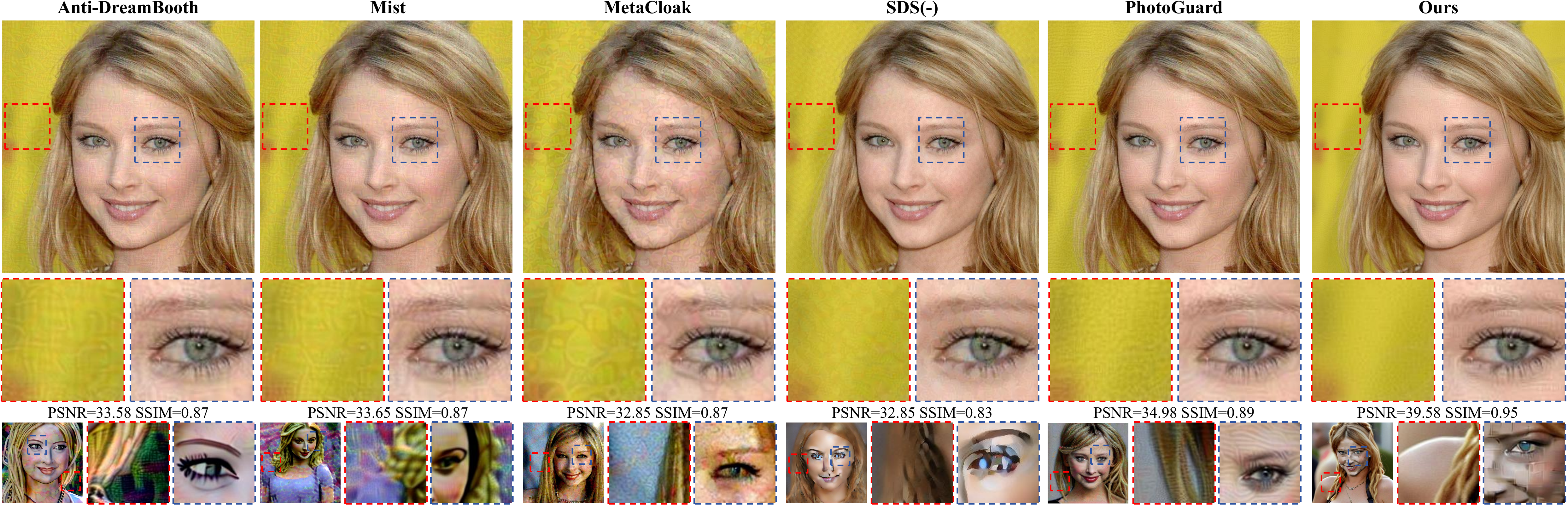}
    \caption{The first row shows protected images from Anti-DreamBooth~\cite{le_etal2023antidreambooth},
    Mist~\cite{zhang2022towards}, MetaCloak~\cite{liu2024metacloak}, SDS(-)~\cite{xue2024toward}, PhotoGuard~\cite{salman2023raising}, and our VCPro. 
    Second and third rows show magnified regions and Textual Inversion results. At $\epsilon$=8/255, our method better balances protection effectiveness and visual quality (higher PSNR/SSIM).
    }
    \label{fig:enter-label}
\end{figure*}

One prevailing direction to mitigate these potential risks is to leverage adversarial attack techniques, transforming original images into adversarial examples, termed ``protected images''. 
These protected images can misguide the personalized diffusion model and deceive its generation process, which resists malicious editing or personalization~\cite{salman2023raising,liang2023adversarial,liang2023mist,le_etal2023antidreambooth,shan2023glaze,liu2024metacloak}.
For instance, AdvDM~\cite{liang2023adversarial} employs adversarial attacks in the inversion stage of Stable Diffusion to safeguard against malicious imitation of the style of a specific artist.
Later, its updated version, Mist~\cite{liang2023mist}, enhances the protection efficacy of protected images by adding a textual loss, extending its application from Textual Inversion to DreamBooth.
Unlike AdvDM and Mist, which focus on art style protection, Anti-DreamBooth~\cite{le_etal2023antidreambooth} undermines DreamBooth model generation quality to enhance privacy by adding adversarial perturbations to human face images before posting online.
Furthermore, MetaCloak~\cite{liu2024metacloak} leverages a meta-learning strategy and data transformations to generate more effective and robust protected images against DreamBooth.

However, these methods primarily focus on preventing the personalization methods from generating high-quality corresponding images, often resulting in noticeable and unacceptable perturbations.
They usually use $11/255$~\cite{le_etal2023antidreambooth,shan2023glaze,liu2024metacloak} or $17/255$~\cite{liang2023mist} as perturbation size in their paper, which is generally unacceptable for owners of face photos, significantly limiting the usability of these methods in real-world applications.
Therefore, we pose the following question in this work:
\textit{How can we find a better trade-off between the visibility of perturbations and protection effectiveness?}

To answer this question, we highlight the sparsity of images under concept protection tasks: The critical information that deserves protection constitutes only a part of the image.
Previous approaches aim to protect the entire image, including background regions and other non-essential information, enhancing the visibility of perturbations.
As shown in Fig.~\ref{fig:enter-label}, the protected images generated by Mist, Anti-DreamBooth, MetaCloak, SDS(-), and PhotoGuard, exhibit noticeable odd textures on the entire image: face, neck, and background, and their final protective effects akin to a special style picture of the target person.
In contrast, we prioritize protecting essential information within an image, utilizing a more stealthy adversarial perturbation.

To this end, we introduce a \underline{V}isual-Friendly \underline{C}oncept \underline{Pro}tection (VCPro) framework to counteract unauthorized concept-driven text-to-image synthesis.
This framework learns selective adversarial perturbations targeting important regions.
Unlike discriminative tasks where important information is class-related and provided by target model gradients, identifying crucial information in generative tasks is challenging.
VCPro utilizes user-provided masks for target concept protection, which can be supplemented by other privacy detection tools for online platforms.
In that way, we propose a regional adversarial loss using spatial information to focus on selected areas. To further enhance visual quality, we apply a Lagrangian multiplier-based solution, shifting from maximizing protection effectiveness to minimizing perceptibility while ensuring effective protection.
Considering human sensitivity to low-frequency changes, we measure perturbation perceptibility in the frequency domain.
Experiments on models like Textual Inversion and DreamBooth validate VCPro's effectiveness. Our approach yields subtler adversarial perturbations compared to baselines like Mist and Anti-DreamBooth,  especially FID, which is reduced from $96.24$ to $27.04$.

Our contributions are summarized as follows:
\begin{itemize}
    \item We point out that the existing image protection methods over-emphasize the final protection effectiveness while neglecting the visual appearance of the protected images.
    \item We propose a visual-friendly concept protection framework that uses regional adversarial loss to protect essential image information. Considering human sensitivity, we measure perturbation perceptibility in the frequency domain and optimize for the smallest feasible perturbations rather than the strongest ones within size constraints.
    \item Experiments demonstrate that our approach can focus on crucial concepts specified by users with lower perceptibility than baselines, achieving a better trade-off between protection effectiveness and perturbation visibility.
\end{itemize}
\section{Related Work}
\noindent \textbf{Personalization of Diffusions Models.}
Personalization for specific concepts (attributions, styles, or objects) has been a long-standing goal in the image generation field.
In text-to-image diffusion models, previous researchers have primarily concentrated on prompt learning and test-time tuning of pre-trained models to generate images based on specific target concepts using special language tokens.
Textual Inversion adjusts text embeddings of a new pseudoword to describe the concept~\cite{gal2022image}.
DreamBooth fine-tunes denoising networks to connect the novel concept and a less commonly used word token~\cite{ruiz2023dreambooth}.
Based on that, more works~\cite{voynov2023p+,zhang2023prospect,kumari2023multi} are proposed to improve controllability and flexibility in processing image visual concepts.
In this paper, we have selected Textual Inversion and DreamBooth as the techniques used by the adversary due to their popularity and representativeness.

\noindent \textbf{Imperceptibility Adversarial Attack.}
Adversarial examples~\cite{szegedy2013intriguing,carlini2017towards,duan2021advdrop,mi2023adversarial} are initially introduced by adding imperceptible noise to original data, fooling classifiers into misclassifying with high confidence.
Recently, more and more researchers aim to improve the imperceptibility of adversarial examples, and they make use of a variety of tools such as perceptual color distance~\cite{zhao2021invertible}, low-frequency spatial constraints~\cite{luo2022frequency}, hybrid attacks in frequency and spatial domain~\cite{jia2022exploring}, and invertible neural networks~\cite{chen2023imperceptible}, etc.
But they are mainly aimed at discriminative tasks such as image classification, where important information in the image can be fed back relatively accurately by the gradient of the target model, whereas we target a diffusion-based generative model whose gradient is also for the whole image.

\noindent \textbf{Adversarial Examples Against Unauthorized Diffusion Generation.}
Unauthorized AI generation poses significant safety risks, driving research into mitigation approaches. While passive defenses focus on detecting synthetic images~\cite{wu2022robust,li2017localization}, adversarial attacks offer promising protection against unauthorized generation~\cite{ruiz2020disrupting, wanganti, wang2022deepfake, zhu2023information}.
Recent works target diffusion models specifically. Photoguard~\cite{salman2023raising} attacks VAE encoders to prevent malicious editing, while AdvDM~\cite{liang2023adversarial} protects artistic styles by maximizing denoising loss. Mist~\cite{liang2023mist} enhances AdvDM with texture loss in latent space. GLAZE~\cite{shan2023glaze} measures art style similarity using pre-trained style-transfer models. Anti-DreamBooth~\cite{le_etal2023antidreambooth} protects privacy from DreamBooth learning, with MetaCloak~\cite{liu2024metacloak} improving robustness via meta-learning. Xue~\etal~\cite{xue2024toward} introduce Score Distillation Sampling (SDS) loss to reduce computational costs.
However, except for GLAZE, existing methods protect entire images with significant noise, degrading user experience and potentially overemphasizing backgrounds while missing critical information. We propose prioritizing limited noise to protect important semantic regions like faces and specific IPs, unlike GLAZE which targets art style.

\begin{figure*}
    \centering
    \includegraphics[width=\linewidth]{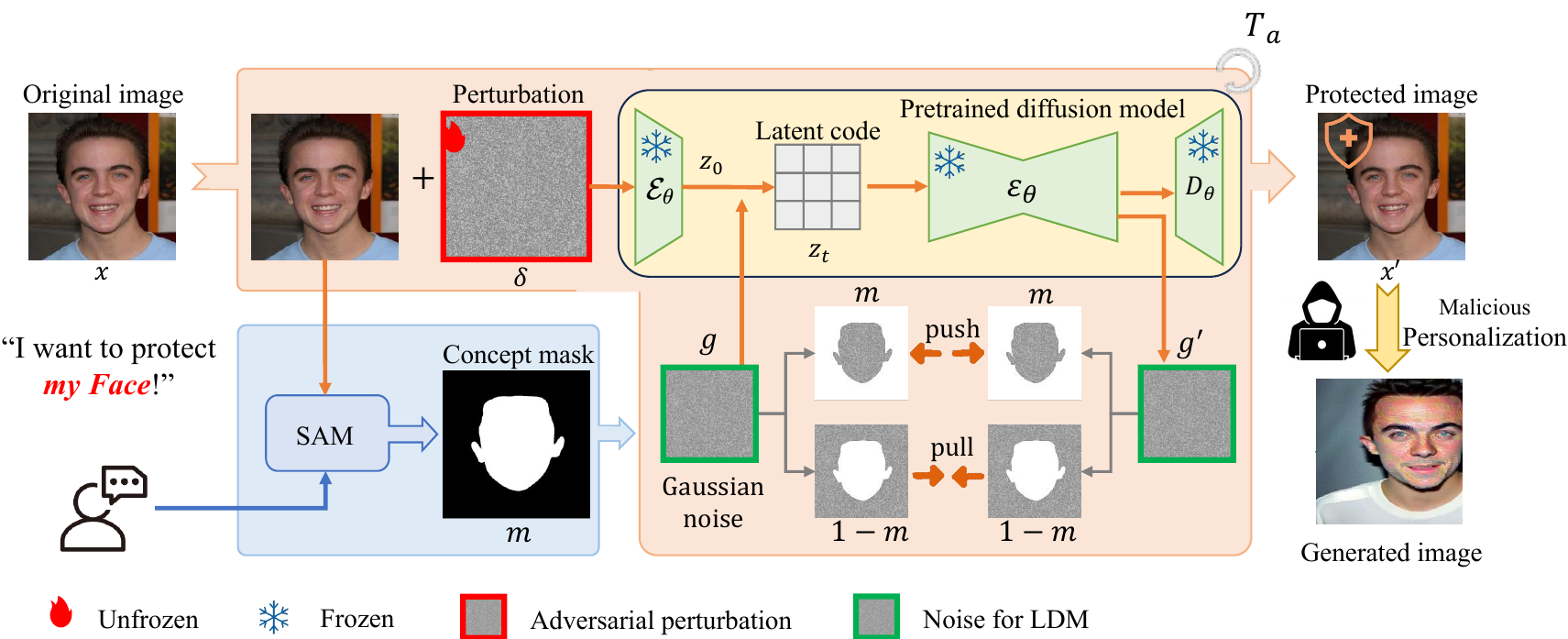}
    \caption{VCPro pipeline. Users create protective masks via SAM or other tools. The masks and images are fed into our protection module (Eq.~\ref{eq:final}), which uses regional adversarial learning to minimize perturbation visibility while maintaining protection.
    }
    \label{fig:pipeline}
\end{figure*}

\section{Preliminaries}
\noindent \textbf{Personalization based on Diffusion.}
Concept-driven personalization customizes generative model outputs to align with specific concepts. Most techniques apply to latent diffusion models (LDMs, parameterized by $\theta$) consisting of image encoder $\mathcal{E}_{\theta}$, decoder $\mathcal{D}_{\theta}$, condition encoder $\tau_{\theta}$, and denoising UNet $\varepsilon_{\theta}$. For image $x$ with latent code $z_0 = \mathcal{E}_{\theta}(x)$, the training objective is:
\begin{equation}
\label{eq:ldm}
    \mathcal{L_{\theta}} := \mathbb{E}_{z \sim \mathcal{E}(x), y, g\sim \mathcal{N}(0,1), t}\left[\left\|{g- \varepsilon_{\theta}\left(z_{t}, t,\tau_{\theta}(y) \right)}\right\|_{2}^{2}\right].
\end{equation}
Textual Inversion learns embeddings $v$ for pseudo tokens $sks$ in prompts like ``a photo of $sks$ [class noun]'' by optimizing:
\begin{equation}
   \arg \min_{v} \mathbb{E}_{z, y, g, t}\left[\left\|g-\varepsilon_{\theta}\left(z_{t}, t, \tau_{\theta}(y)\right)\right\|_{2}^{2}\right].
\end{equation}
DreamBooth fine-tunes LDM parameters on training images and class examples $x^p$ to prevent catastrophic forgetting:
\begin{equation}
\begin{gathered}
    \theta := \arg \min_{\theta} \mathbb{E}_{z, y, g, t} \Bigg[ \left\|g - \varepsilon_{\theta}\left(z_{t}, t, \tau_{\theta}(y)\right)\right\|_{2}^{2} \\
    + \left\|g - \varepsilon_{\theta}\left(z^p_{t}, t^p, \tau_{\theta}(y^p)\right)\right\|_{2}^{2} \Bigg].
\end{gathered}
\end{equation}

\noindent \textbf{Protected Image for Diffusions.}
Protected images add imperceptible adversarial perturbations $\delta$ to original images $x$ to disrupt personalization. The protected image $x'$ is formalized as:
\begin{equation}
    \begin{gathered}
    x^{\prime}:= \arg \max_{x^{\prime}} \mathcal{L_{\theta}}(x^{\prime},y), \\
    \text{s.t.} \quad\left\|x-x^{\prime}\right\| \leq \epsilon,
    \end{gathered}
\end{equation}
where $\epsilon$ limits the perturbation budget. These adversarial attacks create samples difficult to denoise, enhancing LDM optimization challenges for image protection.
\begin{figure*}
    \centering
    \includegraphics[width=\linewidth]{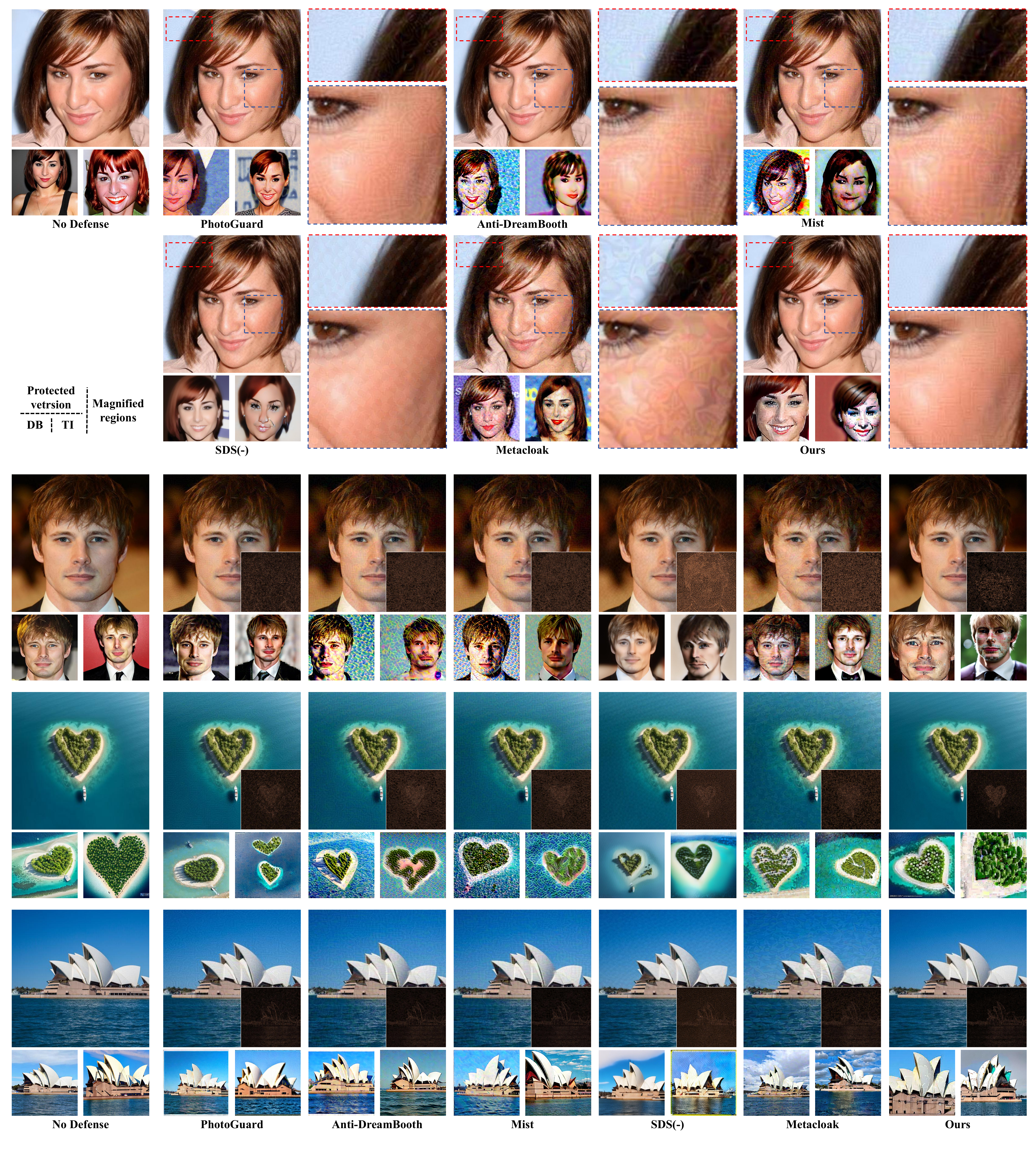}
    \caption{Qualitative comparison ($\epsilon=8/255$, SD v1-4). Rows show original and protected images from baseline methods and ours, with TI/DB results. Perturbations visualized in black-yellow. Please zoom in.}
    \label{fig:case_main}
\end{figure*}
\section{Visual-Friendly Concept Protection}
\subsection{Overview}
Fig.~\ref{fig:pipeline} shows the pipeline of the proposed framework for visual-friendly concept protection.
Accurately describing spatial positions through language can be challenging for users, but it can be precisely achieved using masks.
By leveraging SAM~\cite{Kirillov_2023_ICCV} or other segmentation tools, users can generate a mask $m$ for important concepts within a given image $x$.
The user-provided masks and original images are then collectively fed into the protected image generation module as described in Eq.~(\ref{eq:final}).
In this module, we propose a regional adversarial learning loss to reduce the visibility of protected images through precise protection and a Lagrangian multiplier-based solution to minimize perturbations while maintaining successful protection.

In this section, we start by formulating the regional adversarial learning framework for diffusion models in Sec.~\ref{sec:aeg} and then move on to the solution of the proposed optimization objectives in Sec.~\ref{sec:Lagrangian}. 
The total learning process is shown in Alg.~\ref{alg:algorithm}.

\vspace{-2mm}
\subsection{Formulation}
\label{sec:aeg}

\noindent  \textbf{Regional Adversarial Loss.}
Unlike previous studies, we aim to achieve precise concept protection to reduce the perceptibility of protected images.
We use mask $m$ to indicate the spatial positions of important information in the feature map enabling precise concept protection.
This precise optimization allows prioritized protection of the most critical information in the image with a smaller adversarial perturbation.
The optimization objective of protected images in our method can be defined as:
\begin{equation}
    \begin{gathered}
     x^{\prime}:=\arg \max _{x^{\prime}} \mathcal{L}^{\prime}_\theta\left(x^{\prime},y,m\right),\\
    \text { s.t. } \quad\left\|x-x^{\prime}\right\| \leq \epsilon,\\
    \end{gathered}
    \label{eq:ours_w_mask}
\end{equation}
{and the regional adversarial loss $\mathcal{L}^{\prime}_{\theta}$ combines a ``push'' term for protected regions and a ``pull'' term for non-protected regions.}

\begin{align}
    \mathcal{L}^{\prime}_{\theta} := & \mathbb{E}_{z \sim \mathcal{E}(x), y, g \sim \mathcal{N}(0,1), t}\left[l_{mask}(z_t,y,g,t,m) \right],\label{eq:ours_L} \\
    l_{mask} := & \left\|\Delta \odot m \right\|_{2}^{2} - \left\|\Delta \odot (1 - m) \right\|_{2}^{2}, \nonumber
\end{align}
where $\Delta = g - \varepsilon_{\theta}(z_{t}, t, \tau_{\theta}(y))$.
This loss operates through a balanced mechanism of opposing forces. The push term maximizes the distance between the predicted noise $\varepsilon_{\theta}$ and ground truth noise $g$ in masked regions ($m=1$), effectively disrupting the denoising process for protected concepts. Simultaneously, the pull term minimizes this distance in unmasked regions ($m=0$), preserving visual quality in non-protected areas.

During optimization, these components generate distinct gradient signals: push gradients divert predictions away from ground truth in protected regions, while pull gradients maintain accuracy elsewhere. This dual approach is crucial for effective protection—without the pull component, regions outside the mask cannot provide sufficient gradient feedback, significantly impairing the optimization process. Our ablation studies in Table~\ref{ex:ablation} demonstrate that removing the pull component substantially degrades protection effectiveness.

\begin{algorithm}[tb]
\caption{Visual-Friendly concept protection~(VCPro) framework}
\label{alg:algorithm}
\begin{algorithmic}[1] 
\REQUIRE Image $x$, diffusion model with parameter $\theta$, number of time steps $T$, text condition $y$, attack steps $T_a$, step size $\alpha$, and adversarial perturbation size $\epsilon$\\
\STATE Initialize $x'=x$, $i=0$
\STATE Get mask $m$ by SAM or user-provide
\WHILE{$i<T_{a}$}
\STATE Sample $t \sim [0,T]^n$
\STATE $\delta=\operatorname{Uniform}(-\epsilon, \epsilon)$
\STATE Calculate $\mathcal{L}_{final}(x',y,m)$ by Eq.~(\ref{eq:final})
\STATE $\delta=\delta-\alpha \cdot \operatorname{sign}\left(\nabla_{\delta} \mathcal{L}_{final}(x',y,m)\right)$ 
\STATE $\delta=\max (\min (\delta, \epsilon),-\epsilon) $
\STATE $x' = x'+\delta$
\STATE $x'=\max (\min (x', 255),0) $
\STATE $i=i+1$
\ENDWHILE
\STATE \textbf{return} Protected image $x'$
\end{algorithmic}
\end{algorithm}

\subsection{Lagrangian Multiplier-based Loose Solution}
\label{sec:Lagrangian}
Based on Eq.~\eqref{eq:ours_L}, protected images will destroy the target area as much as possible.
However, for the generation task, the output can still be recognized as synthetic as long as there are some clear signs of protection.
Typical image classification attacking methods~\cite{carlini2017towards,luo2022frequency,chen2023imperceptible}  also present similar viewpoints that minimize the impact on normal visual perception while maintaining the effectiveness of the attack.
Hence, we propose a loosed optimization objective by attempting to find the minimal adversarial perturbation $\delta$ that can attack diffusion models successfully.
For convenience, we set a heuristic method: if $\mathcal{L}^{\prime}_{\theta}>\alpha$, the attack is successful.
The problem can be translated as:

\begin{equation}
\begin{aligned}
\text{minimize} \quad & D(x, x + \delta), \\
\text{such that} \quad & -\mathcal{L}^{\prime}_{\theta}+\alpha \leq 0, \\
& \delta \in [-\epsilon, \epsilon]^n ,\\
& x+\delta \in [0,255]^n,
\label{eq:cw_fomal}
\end{aligned}
\end{equation}
where $D(\cdot)$ is a distance metric.

For ease of solution, we use Lagrangian multiplier method and get the alternative formulation:

\begin{equation}
\begin{aligned}
\text{minimize} \quad & \mathcal{L}_{final}=c \cdot D(x, x + \delta) -\mathcal{L}^{\prime}_{\theta}+\alpha,\\
\text{such that} \quad & \delta \in [-\epsilon, \epsilon]^n , \\
& x+\delta \in [0,255]^n,
\end{aligned}
\label{eq:final}
\end{equation}
where a constant \( c > 0 \) is appropriately selected.
The equivalence of Eq.~(\ref{eq:cw_fomal}) and Eq.~(\ref{eq:final}) can be understood by the existence of a positive constant \( c \) ensuring the best solution for the second formulation aligns with that of the first.

Considering that the human visual system is more sensitive to low-frequency regions~\cite{luo2022frequency}, we use $D(\cdot)$ to limit perturbations into the high-frequency regions.
Specifically, we use discrete wavelet transform (DWT) to transform images from the spatial domain to the frequency domain.
DWT will decompose the image $x$ into one low-frequency and three high-frequency components, i.e., \( x_{ll}, x_{lh}, x_{hl}, x_{hh} \), and inverse DWT (IDWT) uses all four components to reconstruct the image.
\[
x_{ll} = LxL^T, \quad x_{lh} = HxL^T,
\]
\[
x_{hl} = LxH^T, \quad x_{hh} = HxH^T,
\]
where \( L \) and \( H \) are an orthogonal wavelet's low-pass and high-pass filters, respectively. \( x_{ll} \) preserves the low-frequency information of the original image, whereas \( x_{lh}, x_{hl} \) and \( x_{hh} \) are associated with edges and drastic variations.

In this work, we drop the high-frequency components and reconstruct an image with only the low-frequency component as \( \widetilde{x} = \phi(x) \), where
$\phi(x) = L^T x_{ll} L = L^T (LxL^T) L,
\text{and } D(\cdot) = ||\widetilde{x}-\widetilde{x^\prime}||_{2}^{2}$.

\begin{table*}[t]
    \centering
    \small  
    \setlength{\tabcolsep}{4pt}  
    \begin{tabular}{llccccccccccc}
    \toprule
    \multirow{2}{*}{\textbf{Dataset}} &
    \multirow{2}{*}{\textbf{Method}} &
    \multicolumn{3}{c}{\textbf{Visibility}} &
    \multicolumn{2}{c}{\textbf{TI}} &
    \multicolumn{2}{c}{\textbf{DB}} &
    \multicolumn{3}{c}{\textbf{Human Eval}} \\
    \cmidrule(lr){3-5} \cmidrule(lr){6-7} \cmidrule(lr){8-9} \cmidrule(lr){10-12}
    & & \textbf{FID$\downarrow$} & \textbf{SSIM$\uparrow$} & \textbf{PSNR$\uparrow$} & \textbf{LIQE$\downarrow$} & \textbf{CLIP-F$\downarrow$} & \textbf{LIQE$\downarrow$} & \textbf{CLIP-F$\downarrow$} & \textbf{Vis. (L/T/W)} & \textbf{TI$\uparrow$} & \textbf{DB$\uparrow$} \\
    \midrule
    \multirow{7}{*}{\rotatebox{90}{\textbf{VGGFace2}}} 
    & No Defense & -- & -- & -- & 3.61$_{\pm0.98}$ & 0.24$_{\pm0.34}$ & 3.95$_{\pm1.11}$ & 0.38$_{\pm0.25}$ & \textbf{64}/25/11 & -- & -- \\
    & Mist & 105.9$_{\pm9.8}$ & 0.83$_{\pm0.03}$ & 32.5$_{\pm1.1}$ & 1.24$_{\pm0.41}$ & 0.11$_{\pm0.29}$ & \textbf{1.02$_{\pm0.11}$} & 0.25$_{\pm0.28}$ & 4/6/\textbf{90} & 98.3 & 97.2 \\
    & Anti-DB & 96.2$_{\pm10.4}$ & 0.82$_{\pm0.03}$ & 32.4$_{\pm1.1}$ & \textbf{1.18$_{\pm0.36}$} & 0.05$_{\pm0.29}$ & 1.03$_{\pm0.14}$ & 0.26$_{\pm0.28}$ & 3/9/\textbf{88} & 97.4 & \textbf{100.0} \\
    & PhotoGuard & 62.1$_{\pm8.2}$ & 0.84$_{\pm0.03}$ & 33.2$_{\pm1.2}$ & 1.50$_{\pm0.54}$ & 0.13$_{\pm0.34}$ & 1.26$_{\pm0.30}$ & 0.27$_{\pm0.31}$ & 13/19/\textbf{68} & \textbf{98.7} & 84.0 \\
    & SDS(-) & 47.5$_{\pm6.8}$ & 0.81$_{\pm0.03}$ & 32.9$_{\pm0.4}$ & 2.09$_{\pm0.82}$ & 0.07$_{\pm0.37}$ & 2.74$_{\pm0.97}$ & 0.33$_{\pm0.26}$ & 13/16/\textbf{71} & 94.6 & 77.1 \\
    & MetaCloak & 204.3$_{\pm54.5}$ & 0.82$_{\pm0.02}$ & 32.1$_{\pm0.6}$ & 1.86$_{\pm0.62}$ & 0.19$_{\pm0.33}$ & 1.33$_{\pm0.41}$ & 0.25$_{\pm0.32}$ & 2/10/\textbf{88} & 91.0 & 90.4 \\
    & VCPro & \textbf{27.0$_{\pm4.4}$} & \textbf{0.90$_{\pm0.03}$} & \textbf{35.3$_{\pm1.8}$} & 2.31$_{\pm0.78}$ & \textbf{0.03$_{\pm0.32}$} & 2.06$_{\pm0.70}$ & \textbf{0.21$_{\pm0.34}$} & -- & 98.6 & 97.4 \\
    \midrule
    \multirow{7}{*}{\rotatebox{90}{\textbf{CelebA-HQ}}} 
    & No Defense & -- & -- & -- & 4.40$_{\pm0.86}$ & 0.41$_{\pm0.22}$ & 4.84$_{\pm0.40}$ & 0.63$_{\pm0.13}$ & \textbf{77}/19/4 & -- & -- \\
    & Mist & 78.3$_{\pm11.8}$ & 0.86$_{\pm0.03}$ & 33.8$_{\pm0.2}$ & \textbf{1.45$_{\pm0.58}$} & 0.26$_{\pm0.26}$ & \textbf{1.04$_{\pm0.11}$} & 0.44$_{\pm0.21}$ & 5/4/\textbf{91} & 96.3 & \textbf{100.0} \\
    & Anti-DB & 78.5$_{\pm11.7}$ & 0.86$_{\pm0.03}$ & 33.7$_{\pm0.2}$ & 1.81$_{\pm1.20}$ & 0.28$_{\pm0.25}$ & 1.04$_{\pm0.15}$ & 0.44$_{\pm0.21}$ & 0/4/\textbf{96} & 97.0 & 96.5 \\
    & PhotoGuard & 45.2$_{\pm7.2}$ & 0.88$_{\pm0.02}$ & 35.1$_{\pm0.3}$ & 1.81$_{\pm0.59}$ & 0.30$_{\pm0.24}$ & 1.23$_{\pm0.35}$ & 0.48$_{\pm0.17}$ & 2/22/\textbf{76} & 94.3 & 94.3 \\
    & SDS(-) & 34.5$_{\pm6.2}$ & 0.82$_{\pm0.03}$ & 33.0$_{\pm0.3}$ & 2.29$_{\pm0.70}$ & 0.09$_{\pm0.28}$ & 2.92$_{\pm0.74}$ & 0.55$_{\pm0.15}$ & 9/11/\textbf{80} & 98.6 & 84.3 \\
    & MetaCloak & 161.9$_{\pm26.9}$ & 0.86$_{\pm0.03}$ & 33.1$_{\pm0.2}$ & 1.47$_{\pm0.56}$ & 0.25$_{\pm0.26}$ & 1.59$_{\pm0.53}$ & 0.44$_{\pm0.16}$ & 0/4/\textbf{96} & 96.0 & 95.5 \\
    & VCPro & \textbf{16.2$_{\pm2.9}$} & \textbf{0.95$_{\pm0.01}$} & \textbf{39.3$_{\pm0.4}$} & 2.61$_{\pm1.00}$ & \textbf{0.09$_{\pm0.28}$} & 2.62$_{\pm0.95}$ & \textbf{0.44$_{\pm0.19}$} & -- & \textbf{100.0} & 95.7 \\
    \bottomrule
    \end{tabular}
    \caption{Comparison of protection methods on VGGFace2 and CelebA-HQ with $\epsilon=8/255$. Metrics include visibility (FID, SSIM, PSNR), quality degradation (LIQE, CLIP-FACE), and protection efficacy (TI/DB: \% synthetic images detected). Human evaluation visibility shows Lose/Tie/Win rates for VCPro vs. baselines. $\uparrow$/$\downarrow$ indicates higher/lower is better. \textbf{Bold}: best performance.}
    \label{ex:Main-auto}
\end{table*}
\section{Experiments}
This section introduces our experimental settings and qualitative and quantitative experiments to demonstrate our effectiveness in generating protected images with less visual perceptibility while safeguarding key concepts in user-provided images.

\subsection{Experimental Settings}
We validate our method on CelebA-HQ~\cite{karras2017progressive} and VGGFace2~\cite{cao2018vggface2} datasets, randomly selecting 50 identities from each with at least 15 images exceeding 500×500 resolution, following Anti-DreamBooth~\cite{le_etal2023antidreambooth} and MetaCloak~\cite{liu2024metacloak}. Using Stable Diffusion v1-4 with 512×512 resolution, we generate protected images with 720 iterations, step size $\alpha=1/255$, and perturbation size $\epsilon=8/255$. We test protection performance against Textual Inversion~\cite{gal2022image} (learning rate $5 \times 10^{-4}$, 3000 steps) and DreamBooth~\cite{ruiz2023dreambooth} (learning rate $5 \times 10^{-7}$, 1000 steps), comparing with five SOTA baselines: Mist~\cite{liang2023adversarial}, Anti-DreamBooth~\cite{le_etal2023antidreambooth}, PhotoGuard~\cite{salman2023raising}, SDS(-)~\cite{xue2024toward}, and MetaCloak~\cite{liu2024metacloak}. We evaluate visual perception using FID~\cite{heusel2017gans}, SSIM~\cite{wang2004image}, and PSNR between protected and original images, and assess protection effectiveness using LIQE~\cite{zhang2023blind} for full image quality and CLIP-FACE~\cite{liu2024metacloak} for face regional quality evaluation.
LIQE measures image quality on a five-point scale: $c\in C=\{1, 2, 3, 4, 5\} = \{\text{``bad''}, \text{``poor''},
\text{``fair''}, \text{``good''}, \text{``perfect''}\}$.
CLIP-FACE is based on CLIP-IQA for visual quality by considering additional class information. All experiments are conducted on NVIDIA A100 GPU 40GB with parameters $c=0.1$ and $\alpha=0.5$.
We provided detailed experimental settings in \textbf{Supplementary Materials}.
\begin{table*}[t]
    \centering
    \small  
    \setlength{\tabcolsep}{5pt}  
    \begin{tabular}{lccccccc}
    \toprule
    \multirow{2}{*}{\textbf{Method}} &
    \multicolumn{3}{c}{\textbf{Visibility}} &
    \multicolumn{2}{c}{\textbf{TI}}&
    \multicolumn{2}{c}{\textbf{DB}}\\
    \cmidrule(lr){2-4} \cmidrule(lr){5-6} \cmidrule(lr){7-8}
    & \textbf{FID$\downarrow$} & \textbf{SSIM$\uparrow$} & \textbf{PSNR$\uparrow$} & \textbf{LIQE$\downarrow$} & \textbf{CLIP-F$\downarrow$} & \textbf{LIQE$\downarrow$} & \textbf{CLIP-F$\downarrow$}\\
    \midrule
    No Defense & -- & -- & -- & 3.61$_{\pm0.98}$ & 0.24$_{\pm0.34}$ & 3.95$_{\pm1.11}$ & 0.38$_{\pm0.25}$ \\
    \midrule
    Anti-DB & 96.2$_{\pm10.4}$ & 0.82$_{\pm0.03}$ & 32.4$_{\pm1.1}$ & \textbf{1.18$_{\pm0.36}$} & 0.05$_{\pm0.29}$ & \textbf{1.03$_{\pm0.14}$} & 0.26$_{\pm0.28}$\\
    Anti-DB+RAL & 68.2$_{\pm8.6}$ & 0.83$_{\pm0.03}$ & 32.9$_{\pm1.3}$ & 1.62$_{\pm0.57}$ & 0.03$_{\pm0.31}$ & 1.53$_{\pm0.53}$ & 0.18$_{\pm0.31}$\\
    Anti-DB+RAL (Small Mask) & 58.4$_{\pm9.3}$ & 0.84$_{\pm0.03}$ & 33.1$_{\pm1.3}$ & 2.11$_{\pm0.82}$ & 0.05$_{\pm0.33}$ & 2.21$_{\pm0.66}$ & 0.27$_{\pm0.32}$\\
    Anti-DB+LMLS & 30.1$_{\pm5.2}$ & 0.90$_{\pm0.03}$ & 35.0$_{\pm1.8}$ & 2.52$_{\pm0.82}$ & 0.17$_{\pm0.34}$ & 2.16$_{\pm0.65}$ & 0.30$_{\pm0.30}$\\
    \midrule
    VCPro (w/o Pull) & \textbf{17.2$_{\pm3.1}$} & \textbf{0.94$_{\pm0.03}$} & \textbf{36.8$_{\pm2.3}$} & 2.74$_{\pm0.82}$ & 0.14$_{\pm0.33}$ & 2.55$_{\pm0.92}$ & 0.35$_{\pm0.32}$\\
    VCPro & 27.0$_{\pm4.4}$ & 0.90$_{\pm0.03}$ & 35.3$_{\pm1.8}$ & 2.31$_{\pm0.78}$ & \textbf{0.03$_{\pm0.32}$} & 2.06$_{\pm0.70}$ & \textbf{0.21$_{\pm0.34}$}\\
    \midrule
    Anti-DB+Input-Mask & 41.3$_{\pm8.8}$ & 0.90$_{\pm0.03}$ & 35.2$_{\pm2.0}$ & 2.05$_{\pm0.76}$ & 0.10$_{\pm0.32}$ & 2.03$_{\pm0.80}$ & 0.23$_{\pm0.33}$\\
    VCPro+Input-Mask & 18.6$_{\pm3.2}$ & 0.93$_{\pm0.03}$ & 36.6$_{\pm2.3}$ & 2.41$_{\pm0.78}$ & 0.07$_{\pm0.34}$ & 2.68$_{\pm1.05}$ & 0.31$_{\pm0.32}$\\
    \bottomrule
    \end{tabular}
    \caption{Ablation study on VGGFace2 with $\epsilon=8/255$. Input-Mask constrains perturbations within the mask region on input images (vs. our method which constrains the optimization objective, allowing perturbations across the entire image). $\uparrow$/$\downarrow$ indicates higher/lower is better. \textbf{Bold}: best performance.}
    \label{ex:ablation}
\end{table*}
\subsection{Main Results}
\label{sec:mr}
\noindent \textbf{Quantitative Results.}
Table~\ref{ex:Main-auto} shows automatic and human evaluation results for protected image visibility and protection effectiveness.
For visibility metrics, our method generates protected images most faithful to originals with stable performance (lowest FID variance). Compared to Anti-DreamBooth baseline, we achieve up to 69.20 FID reduction, 0.09 SSIM improvement, and 5.58 PSNR increase. Our method significantly outperforms current SOTA methods PhotoGuard and SDS (-) in image quality, though SDS (-) achieves low FID by introducing noticeable brightness increases that hurt SSIM performance.

For protection effectiveness, we achieve significant quality degradation in full-image metric LIQE (from ``good/perfect'' to ``poor'') and excel in face region quality metric CLIP-FACE. Both metrics demonstrate reduced visibility against perturbations while maintaining strong protection.

\noindent \textbf{Qualitative Results.}
Fig.~\ref{fig:case_main} shows protected images and effects on identity information and landscapes under Textual Inversion and DreamBooth. Adversarial perturbations are visualized using [0,1] normalization with colormaps.
PhotoGuard, Mist, Anti-DreamBooth, and MetaCloak add obvious strange textures throughout images. SDS(-) causes excessive blurriness with limited DreamBooth protection, adding circular blob artifacts. Our approach achieves subtler perturbations while ensuring protective effectiveness.
Our method prevents personalization methods from generating high-fidelity results. For faces, landmarks, and buildings, our perturbations are less noticeable (especially in backgrounds) while effectively distorting crucial textures and features. This prevents unauthorized use of personal or copyrighted material while maintaining better quality balance.

Textual Inversion protection is easier than DreamBooth since DreamBooth fine-tunes most Stable Diffusion parameters while Textual Inversion only adds word embeddings, as supported by Table~\ref{ex:Main-auto}.

\noindent \textbf{User Study.}
We evaluate VCPro against five methods: PhotoGuard~\cite{salman2023raising}, Mist~\cite{liang2023mist}, Anti-DreamBooth~\cite{le_etal2023antidreambooth}, SDS(-)~\cite{xue2024toward}, and MetaCloak~\cite{liu2024metacloak}.
50 participants (58\% male, 42\% female, ages 18-55, mean: 24.5) with social media proficiency participated.
We randomly sampled protected and generated images from 100 identities in VGGFace2 and CelebA-HQ plus six non-face groups.
The study includes three parts: perturbation visibility (2064 valid votes), Textual Inversion protection (1996 valid votes), and DreamBooth protection (1990 valid votes).

\noindent \textbf{User Study I.}  
Participants compared VCPro-protected images with other methods' results, voting on visual quality (``A wins'', ``tie'', or ``B wins''). 
VCPro received 68\%-96\% of votes against competing methods with Kendall coefficient of 0.71 ($p<0.05$), indicating substantial inter-rater agreement.

\noindent \textbf{User Study II.}  
Participants determined whether images generated via Textual Inversion from protected images were synthetic.
All methods achieved $\ge$ 91\% synthetic recognition, with VCPro showing most effective protection. Cohen's Kappa = 0.88 ($p<0.05$) indicates strong participant agreement.

\noindent \textbf{User Study III.}  
For DreamBooth protection evaluation, participants recognized 97.01\% protective success rate for VCPro.
Cohen's Kappa = 0.88 ($p<0.05$) confirms study consistency.

\begin{table*}[]
    \centering 

    \resizebox{0.8\textwidth}{!}{

    \begin{tabular}{ c   ccc  cc  cc}
    \toprule
    
    \multirow{2}{*}{\textbf{Budget}} &
    \multicolumn{3}{c}{\textbf{Visibility of Perturbations}} &
    \multicolumn{2}{c}{\textbf{Textual Inversion}}&
     \multicolumn{2}{c}{\textbf{DreamBooth}}\\
     ~&\textbf{ FID~$\downarrow$ }&\textbf{ SSIM~$\uparrow$ }& \textbf{PSNR~$\uparrow$} &  \textbf{LIQE~$\downarrow$} &\textbf{CLIP-FACE~$\downarrow$}&
     \textbf{LIQE~$\downarrow$} &\textbf{CLIP-FACE~$\downarrow$}\\
     \midrule
   0& - &  -& - & 3.61$_{\pm0.98}$ & \hphantom{0}0.24$_{\pm0.34}$& 3.95$_{\pm1.11}$ & 0.38$_{\pm0.25}$\\
    4 &\textbf{22.98\boldmath{$_{\pm4.08}$}}&\textbf{0.93\boldmath{$_{\pm0.03}$}}&\textbf{36.45\boldmath{$_{\pm2.21}$}} &2.76$_{\pm0.77}$&-0.02$_{\pm0.27}$&2.33$_{\pm0.70}$&0.25$_{\pm0.31}$ \\
   8 & 27.04$_{\pm4.38}$& 0.90$_{\pm0.03}$ & 35.33$_{\pm1.79}$ &  2.31$_{\pm0.78}$&\hphantom{0}0.03$_{\pm0.32}$ &2.06$_{\pm0.70}$&0.21$_{\pm0.34}$\\
  16 &48.67$_{\pm9.02}$&0.82$_{\pm0.03}$&32.63$_{\pm1.06}$ &\textbf{1.86}\boldmath{$_{\pm0.80}$}&\textbf{-0.02}\boldmath{$_{\pm0.23}$} &\textbf{1.81}\boldmath{$_{\pm0.53}$}&\textbf{0.19}\boldmath{$_{\pm0.32}$}\\
    \bottomrule
    \end{tabular}
    }  \vspace{-2mm}   \caption{The influence of different perturbation size $\epsilon$ on VGGFace2 datasets.
     ↑ means the higher the better, and vice versa. The best result in each column is in bold. }
     \vspace{-4mm}
    \label{ex:eps}
\end{table*}

\subsection{Ablation Study}
\label{sec:ablation}
As shown in Table~\ref{ex:ablation}, we conduct ablation experiments on two VCPro modules: Regional Adversarial Loss (RAL) and Lagrangian Multiplier-based Loose Solution (LMLS), using Anti-DreamBooth as baseline.
Anti-DreamBooth+RAL reduces perturbation visibility with protection concentrated in mask areas, resulting in slight LIQE decrease but stable CLIP-FACE performance. Smaller masks further reduce perturbation visibility but limit protection range.
Anti-DreamBooth+LMLS reduces noise visibility and preserves high-frequency components, but final protection creates streaky textures across the entire image. This confirms that adversarial perturbations in specific spatial/frequency domains directly affect corresponding protection domains.
VCPro without pull-loss (VCPro (w/o Pull)) reduces training gradient feedback for protection while maintaining perturbation visibility constraints, decreasing both visibility and protection effectiveness.
Comparing direct mask-constrained updates (Input-Mask) with our loss-guided approach shows both achieve regional protection. Direct masking improves pixel-level metrics (SSIM/PSNR) but makes perturbations more obvious with high FID. Our combined approach achieves better visual perception while maintaining target protection.

\paragraph{Perturbation Size Influence.}
Perturbation size $\epsilon$ controls the maximum allowable change in pixel values of adversarial perturbation.
As shown in Table~\ref{ex:eps}, different levels of adversarial perturbation size noticeably influence the protection outcomes: Compared to low $\epsilon$, large $\epsilon$ presents worse invisibility of adversarial perturbations while the more obvious protection effects.
For Textual Inversion, when the perturbation size is low, the image preserves the facial area with alterations in facial texture and feature distribution.
Upon reaching a perturbation size of 16/255, the facial areas experience complete degradation.
The results under DreamBooth and quantitative experiment show a similar trend.
Compared with DreamBooth, Textual Inversion is easier to achieve concept protection.

\textbf{Qualitative ablation experiments}, \textbf{hyper-parameter (training iterations, $c$, $\alpha$)}, analysis of adversary settings, and \textbf{frequency domain analysis} please see supplementary materials.

\section{Conclusion}
In this paper, we show that existing approaches utilizing adversarial perturbations to safeguard images from malicious personalization often overemphasize the final protection effectiveness, resulting in more noticeable perturbations.
To mitigate this problem, on the one hand, we protect the important concept regions rather than the full images in previous works, leveraging the sparse nature of images and designing a user-specified image concepts protection framework.
On the other hand, we change the optimization objective from generating the most protective adversarial perturbation to generating the least perceptible adversarial perturbation that exactly achieves the required protective effect.
Quantitative and qualitative experiments demonstrate that we can protect important user-specified concepts and greatly reduce the degree of naked-eye visibility of adversarial perturbations.

\noindent \textbf{Future Works and Limitations.}
Our future efforts will focus on finding efficient methods to produce protected images swiftly while maintaining high visual quality. By doing so, we aim to significantly enhance the overall user experience and ensure that our solutions meet the highest standards of both functionality and aesthetics.

\noindent \textbf{Ethical Considerations.}
VCPro empowers individuals against AI content power imbalances. Technical measures complement limited legal protections.
Dual-use potential—adversaries could theoretically exploit VCPro to evade accountability. However, the primary threat today is unauthorized personalization with limited user consent, and empowering individuals is essential given current power asymmetries.
\section*{Acknowledgments}
This work was partly supported by the National Natural Science Foundation of China under No. 62572458, and the Innovation Funding of ICT, CAS under Grant No. E561160.

\bibliography{aaai2026}

\clearpage
\appendix
\setcounter{page}{1}

\twocolumn[{
\maketitle
\begin{center}
{\LARGE \textbf{Appendix}}\\
\vspace{0.5cm}
{\Large Visual-Friendly Concept Protection via Selective Adversarial Perturbations}
\vspace{1cm}
\end{center}
}]
\setcounter{figure}{3}
\setcounter{table}{2}
\section{Detailed Experimental Settings}
\textbf{Datasets.} 
We used the CelebA-HQ~\cite{karras2017progressive} and VGGFace2~\cite{cao2018vggface2} datasets, following Anti-DreamBooth~\cite{le_etal2023antidreambooth} and MetaCloak~\cite{liu2024metacloak}, to validate the effectiveness of our method in privacy protection.
CelebA-HQ is an enhanced version of the original CelebA dataset consisting of $30,000$ celebrity face images.
VGGFace2 is a large-scale dataset with over $3.3$ million face images from $9,131$ unique identities.
We randomly selected 50 identities from each dataset, ensuring that each had at least 15 images with a resolution exceeding 500 × 500 pixels.
We also provide other non-face cases~(artwork and landscape) in the paper.

\noindent \textbf{Training details.}
The availability of Stable Diffusion pre-trained weights on Hugging Face~\footnote{https://huggingface.co/} has significantly advanced research within the community.
Here, our experiments primarily focus on the Stable Diffusion v1-4 version.
For image pre-processing, we center crop and resize images to a resolution $512 \times 512$.
During protected image generation, we use $720$ iterations, step size $\alpha$ is $1/255$, and adversarial perturbation size $\epsilon$ is $8/255$.
We test the protection performance on two types of SOTA personalization methods, i.e., Textual Inversion (TI)~\cite{gal2022image}, and DreamBooth (DB)~\cite{ruiz2023dreambooth}.
During the training of the Textual Inversion, the constant learning rate is set to $5 \times 10^{-4}$ with $3000$ optimization steps and batch size $1$. 
In DreamBooth, the constant learning rate is set to $5 \times 10^{-7}$ with $1000$ optimization steps and batch size is $2$. 
We use a guidance scale of $7.5$ and $50$ denoising steps at the generation phase with a testing prompt.
We set $c=0.1$ and $\alpha=0.5$ in practice.
All experiments are complete on NVIDIA A100 GPU 40GB.

\noindent \textbf{Baselines.}
We compare our method with five SOTA baselines, i.e., Mist~\cite{liang2023adversarial}, Anti-DreamBooth (referred to as Anti-DB)~\cite{le_etal2023antidreambooth}, PhotoGuard~\cite{salman2023raising}, SDS(-)~\cite{xue2024toward} and MetaCloak~\cite{liu2024metacloak}.
In implementations, we set $\epsilon = 8/255$ as the $\ell_\infty$ perturbation size, $\alpha =1/255$ as the step size for fair comparison.
\begin{itemize}
\item Mist aims to maximize the training loss of Latent Diffusion Models (LDMs) while minimizing a textual loss between the protected image and a dummy image in the image encoder layer during protected image generation. In our implementation, following the original paper~\cite{liang2023mist}, we set the loss weight for the textual loss to $1 \times 10^{-4}$ and run 720 iterations during protected image generation.
\item Anti-DreamBooth focuses on maximizing the training loss of LDMs to generate protected images. We use the most visually imperceptible variant, FSMG, which employs a DreamBooth model trained on original images for protected data generation. The adversarial iterations were set to 720.
\item PhotoGuard targets diffusion-based inpainting/editing methods and offers two variants: one that attacks the VAE-encoder of LDMs, and another that attacks the final image outputs. We select the VAE-encoder variant due to its superior visual quality and set the training iterations to 200, as recommended in the original paper~\cite{salman2023raising}.
\item SDS(-) minimizes the SDS loss~\cite{poole2022dreamfusion} to optimize adversarial data, achieving the best visual quality among all versions presented in its original paper~\cite{xue2024toward}. We set the training iterations to 100, following~\cite{xue2024toward}.
\item MetaCloak utilizes meta-learning to enhance the robustness of adversarial data against various transformations. We set the number of surrogate models to 5, the unrolling number to 1, the sample batch size to 1, and the training iterations to 4000, based on the recommendations in~\cite{liu2024metacloak}.
\end{itemize}
\noindent \textbf{Metrics.}
To measure the visual perception of protected data, we calculate the FID~\cite{heusel2017gans}, SSIM~\cite{wang2004image}, and PSNR between protected and original images.
The goal of the protection is to reduce the visual quality of generated images when adversaries use Textual Inversion and DreamBooth for personalization generation.
So we use a full image quality evaluation metric LIQE~\cite{zhang2023blind} and propose a face regional quality evaluation metric CLIP-FACE~\cite{liu2024metacloak} as automatic evaluation metrics to assess protection effectiveness.
LIQE is an advanced blind image quality assessment tool that measures image quality on a five-point scale: $c\in C=\{1, 2, 3, 4, 5\} = \{\text{``bad''}, \text{``poor''},
\text{``fair''}, \text{``good''}, \text{``perfect''}\}$.
CLIP-FACE is based on CLIP-IQA~\cite{wang2023exploring} for visual quality by considering additional class information.
Specifically, we calculate the CLIP score difference between ``good face'' and ``bad face''.
Considering the large number of images generated by Textual Inversion or DreamBooth, fine-grained labeling of facial regions in every image and then evaluating them is expensive.
CLIP-FACE uses the multimodal understanding capability of CLIP and can focus on the facial regions in a coarse-grained manner.
To assess the efficacy of the protection mechanism, we generate $16$ images for each trained Textual Inversion model and DreamBooth model with testing prompt ``a photo of $sks$ person'' or ``a photo of $sks$ object''.
As image quality assessment remains a challenging problem, particularly with the lack of precise concept-specific automated metrics, we also provide human evaluation to assess the visual perception of protected data and the efficacy of safeguarding the target concepts.
\section{Ablation Study}
\begin{figure*}
    \centering
    \includegraphics[width=\linewidth]{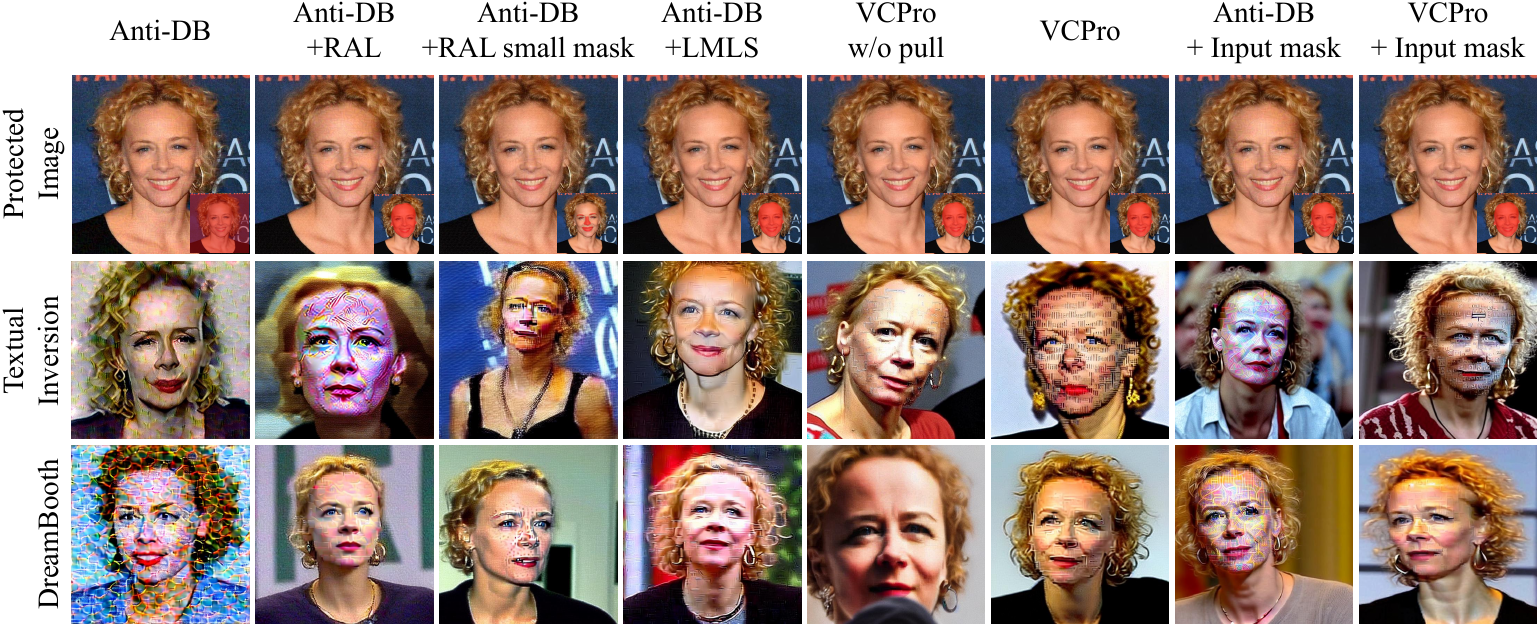}
    \vspace{-4mm}
     \caption{The ablation experiments of different mask sizes and modules. Anti-DreamBooth (Anti-DB) is our baseline. The red region indicates the important region. Here Input-mask means optimizing adversarial perturbations by directly restricting the mask range. Zoom in for a better view.}
     \vspace{-3mm}
    \label{fig:case_ablation}
\end{figure*}

\begin{figure}
    \centering
    \includegraphics[width=\linewidth]{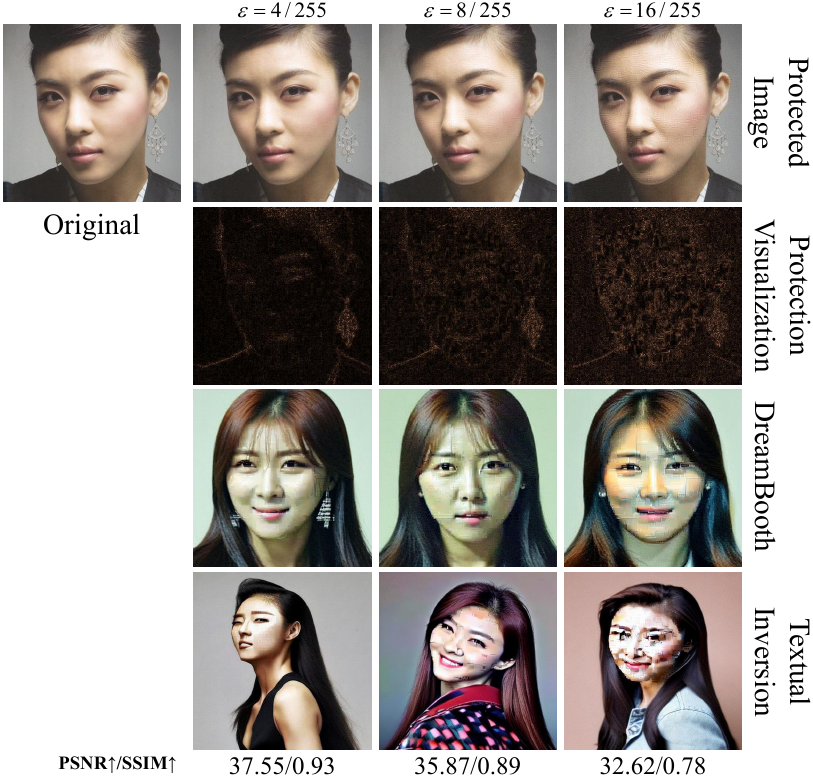}
    \vspace{-4mm}
    \caption{The influence of different $\epsilon$ in VCPro. Numbers under images show the [PSNR$\uparrow$, SSIM$\uparrow$] of each generated protected image. Zoom in for a better view.}
    \label{fig:case_eps}
    \vspace{-4mm}
\end{figure}
As shown in Fig.~\ref{fig:case_ablation} and Table~\ref{ex:ablation}, we conducted ablation experiments of two modules in VCPro: Regional Adversarial Loss~(RAL) and Lagrangian Multiplier-based Loose Solution~(LMLS). Anti-DreamBooth is our baseline.

Compared with Anti-DreamBooth, we find that Anti-DreamBooth+RAL reduces the visibility of the perturbations, while the protection effect is concentrated in the mask area, resulting in a slight decrease in the global image quality assessment LIQE but stable in CLIP-FACE.
At the same time, we also consider the effect of different mask ranges, smaller masks will further reduce the visibility of the perturbations, and the protection range is further reduced.

LMLS shows a similar phenomenon where Anti-DreamBooth+LMLS reduces the visibility of the noise and tends to leave the high-frequency part, while the final protection effect also leaves a streaky texture across the whole image.
Consequently, the adversarial perturbation added to a specific space and frequency domain will directly affect the protection effect in the corresponding space and frequency domains.

Moreover, only push-loss without pull-loss in RAL (VCPro (w/o Pull) in Fig.~\ref{fig:case_ablation} and Table~\ref{ex:ablation}) will reduce the training loss gradient feedback about the protection effect, while the loss that constrains the visibility of the perturbations is not affected resulting in decreased visibility of the perturbations and the protection effect.

We also compare the difference between directly protected images updated within the mask range (Input-Mask) and us adding guidance at the loss side, both can achieve the protection of the specified region.
When adding the mask constraint directly at the protected images, it improves pixel-level global metrics such as SSIM/PSNR.
However, it makes the noise of the protected target on the original image more obvious than us and FID is still high. 
We can achieve a better visual perception by combining the two, and at the same time, achieve the protection of the target.

\section{Perturbation Size Influence}
Perturbation size $\epsilon$ controls the maximum allowable change in pixel values of adversarial perturbation.
As shown in Fig.~\ref{fig:case_eps} and Table~\ref{ex:eps}, different levels of adversarial perturbation size noticeably influence the protection outcomes: Compared to low $\epsilon$, large $\epsilon$ presents worse invisibility of adversarial perturbations while the more obvious protection effects.

For Textual Inversion, when the perturbation size is low, the image preserves the facial area with alterations in facial texture and feature distribution.
Upon reaching a perturbation size of 16/255, the facial areas experience complete degradation.
The results under DreamBooth and quantitative experiment show a similar trend.
Compared with DreamBooth, Textual Inversion is easier to achieve concept protection.

We visualize adversarial perturbations by normalizing all perturbations to the range [0,1] and applying a colormap. An interesting pattern emerges: as $\epsilon$ increases, perturbations spread from the face to the background. This diffusion occurs because, when perturbation size is constrained, the perturbations are concentrated in areas that most effectively enhance the protective effect.

\begin{table*}[]
    \centering 

    \resizebox{0.8\textwidth}{!}{

    \begin{tabular}{ c   ccc  cc  cc}
    \toprule
    
    \multirow{2}{*}{\textbf{Budget}} &
    \multicolumn{3}{c}{\textbf{Visibility of Perturbations}} &
    \multicolumn{2}{c}{\textbf{Textual Inversion}}&
     \multicolumn{2}{c}{\textbf{DreamBooth}}\\
     ~&\textbf{ FID~$\downarrow$ }&\textbf{ SSIM~$\uparrow$ }& \textbf{PSNR~$\uparrow$} &  \textbf{LIQE~$\downarrow$} &\textbf{CLIP-FACE~$\downarrow$}&
     \textbf{LIQE~$\downarrow$} &\textbf{CLIP-FACE~$\downarrow$}\\
     \midrule
   0& - &  -& - & 3.61$_{\pm0.98}$ & \hphantom{0}0.24$_{\pm0.34}$& 3.95$_{\pm1.11}$ & 0.38$_{\pm0.25}$\\
    4 &\textbf{22.98\boldmath{$_{\pm4.08}$}}&\textbf{0.93\boldmath{$_{\pm0.03}$}}&\textbf{36.45\boldmath{$_{\pm2.21}$}} &2.76$_{\pm0.77}$&-0.02$_{\pm0.27}$&2.33$_{\pm0.70}$&0.25$_{\pm0.31}$ \\
   8 & 27.04$_{\pm4.38}$& 0.90$_{\pm0.03}$ & 35.33$_{\pm1.79}$ &  2.31$_{\pm0.78}$&\hphantom{0}0.03$_{\pm0.32}$ &2.06$_{\pm0.70}$&0.21$_{\pm0.34}$\\
  16 &48.67$_{\pm9.02}$&0.82$_{\pm0.03}$&32.63$_{\pm1.06}$ &\textbf{1.86}\boldmath{$_{\pm0.80}$}&\textbf{-0.02}\boldmath{$_{\pm0.23}$} &\textbf{1.81}\boldmath{$_{\pm0.53}$}&\textbf{0.19}\boldmath{$_{\pm0.32}$}\\
    \bottomrule
    \end{tabular}
    }  \vspace{-2mm}   \caption{The influence of different perturbation size $\epsilon$ on VGGFace2 datasets.
     ↑ means the higher the better, and vice versa. The best result in each column is in bold. }
     \vspace{-4mm}
    \label{ex:eps}
\end{table*}

\section{Hyper-parameter Analysis}
Here we provide quantitative results to study the influence of hyper-parameters: training iteration, $c$, and $\alpha$.
We also provide qualitative results in supplementary materials.

\begin{figure}
    \centering
    \includegraphics[width=\linewidth]{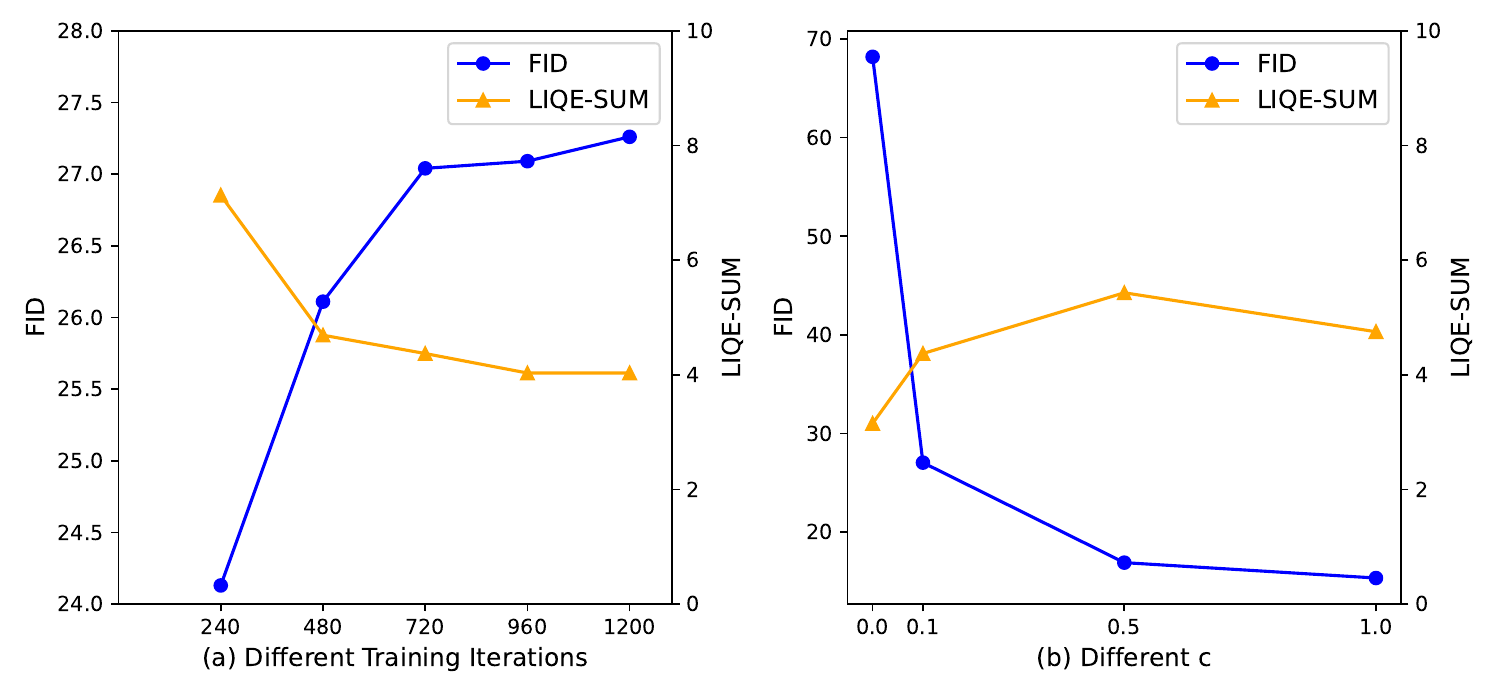}
    \vspace{-8mm}
    \caption{Sensitivity analysis of hyper-parameters training iterations and $c$. FID is calculated between adversarial data and corresponding clean data; here LIQE-SUM is the sum LIQE of Textual Inversion and DreamBooth, which are all the smaller the better. The dataset is VGGface2 and $\epsilon=8/255$.}
    \label{fig:hyp}\vspace{-4mm}
\end{figure}

\begin{figure*}
    \centering
    \includegraphics[width=\linewidth]{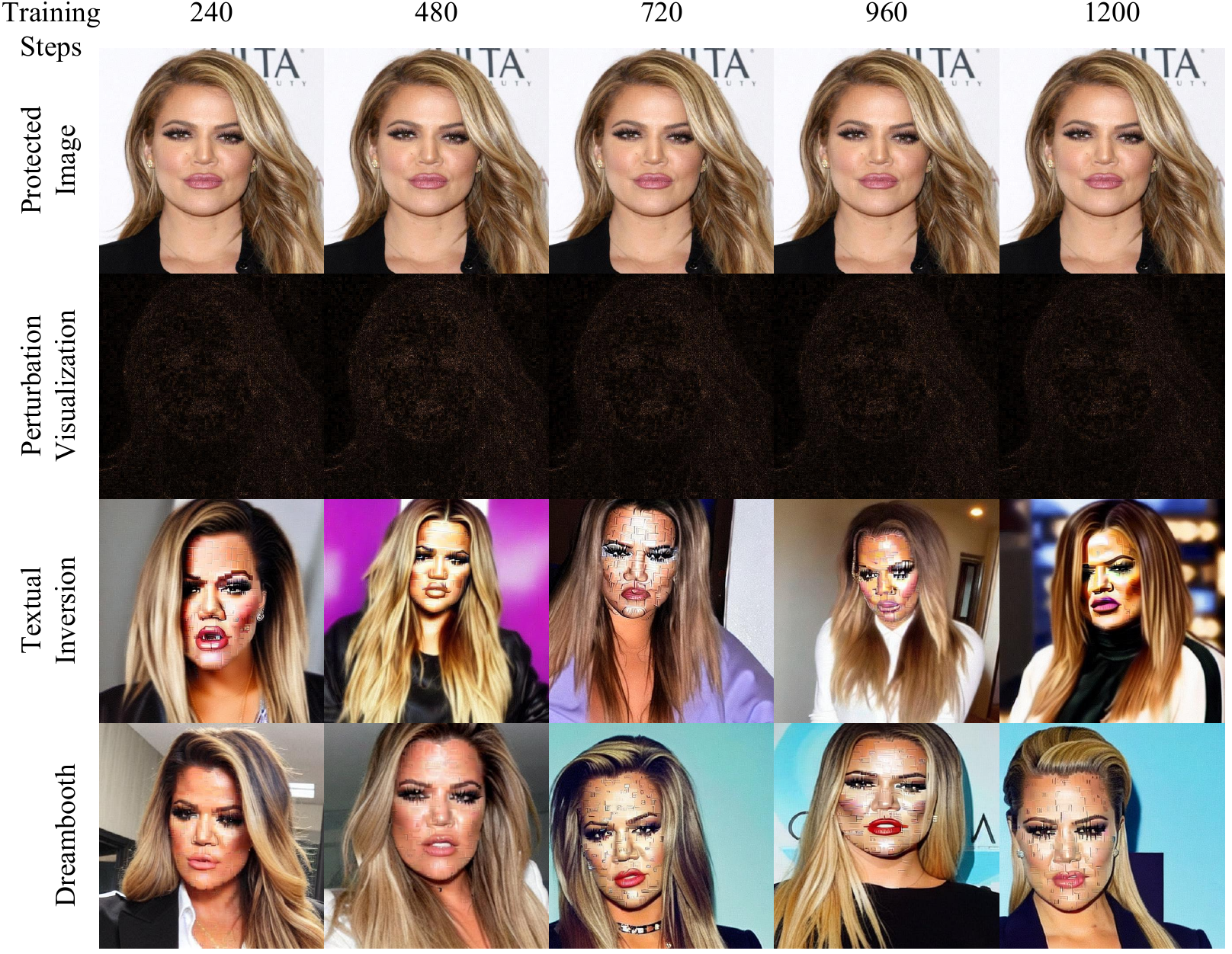}
    \caption{Qualitative results for different training steps. For generating protected images, we use an $\epsilon$ value of 8/255 with Stable Diffusion v1-4. For better detail, please zoom in.}
    \label{fig:case_training_steps}
\end{figure*}

\noindent \textbf{Sensitivity Analysis of Training Iteration.}
Usually, the larger the training steps, the larger the value of $\mathcal{L}^{\prime}_{\theta}$ at different time steps, and the more successful the protection is.
As shown in Fig.~\ref{fig:hyp}~(a), when increasing the number of training iterations, FID will first improve rapidly and then rise slowly, and the overall trend is that the longer the training iterations, the worse the quality of the protected image.
Whereas LIQE will fall rapidly and then level off, the protection effect will improve and then keep stable.
{
Finally, we selected 720 iterations based on the inflection point observed in Fig.~\ref{fig:hyp}~(a). At this point, we achieve substantial protection effectiveness (LIQE-SUM 4.37) while maintaining reasonable image quality (FID  27.04).
Fig.~\ref{fig:case_training_steps} also shows similar pheonmentions. 
Increasing iterations beyond 720 provides only marginal improvements in protection while continuing to degrade image quality.
}

\begin{figure*}
    \centering
    \includegraphics[width=\linewidth]{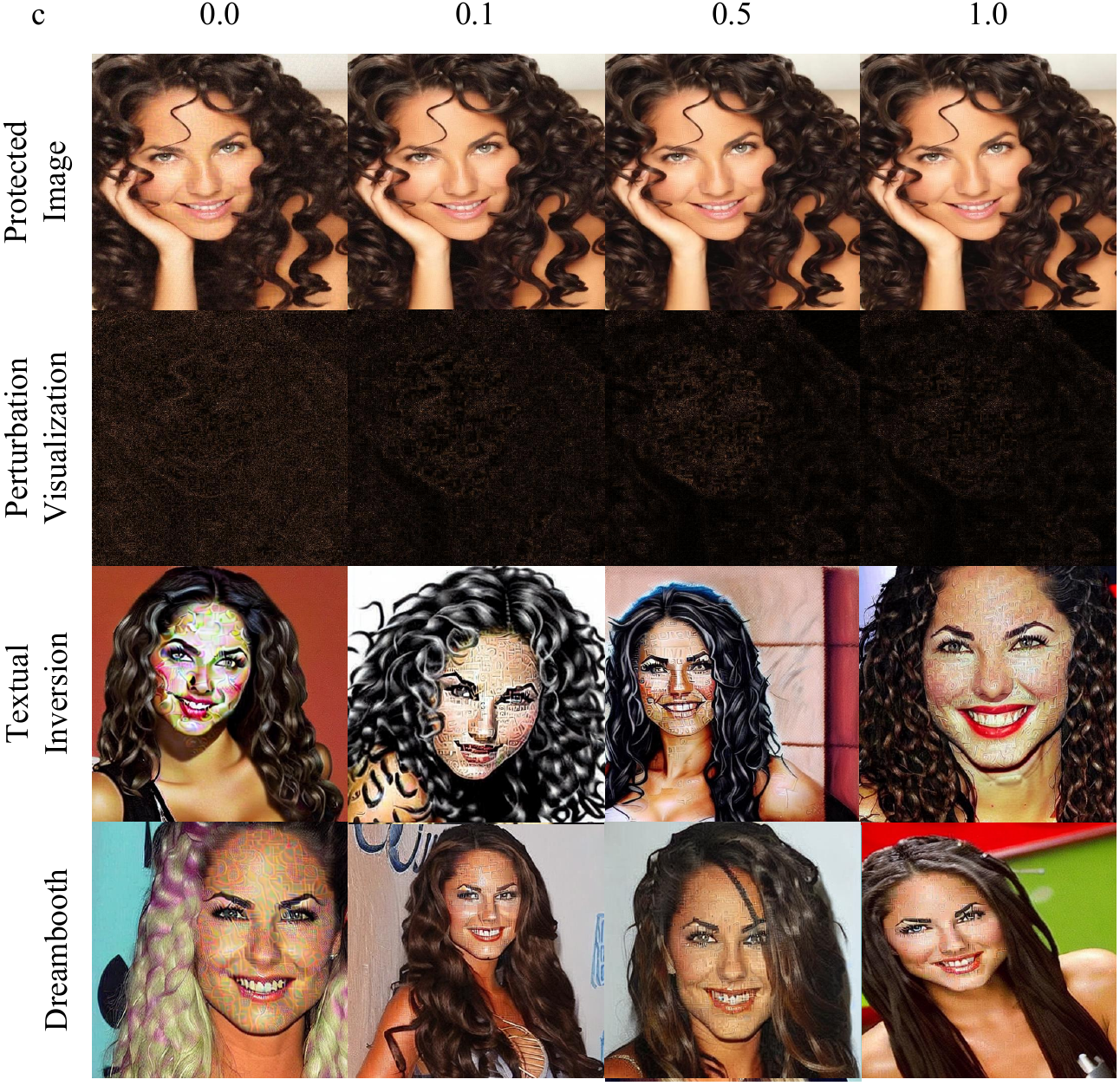}
    \caption{Qualitative results for different $c$. For generating protected images, we use an $\epsilon$ value of 8/255 with Stable Diffusion v1-4. For better detail, please zoom in.}
    \label{fig:case_c}
\end{figure*}

\noindent \textbf{Sensitivity Analysis of $c$.}
{
$c$ represents stronger constraints on perturbation visibility.
As shown in Fig.~\ref{fig:hyp}~(b), increasing parameter $c$ reduces the FID score from 68.18 at $c=0.0$ to 15.33 at $c=1.0$, indicating significant improvements in visual quality of protected images. Meanwhile, protection effectiveness (LIQE-SUM) increases from ~3.15 at $c=0.0$ to ~5.43 at $c=0.5$, then slightly decreases to ~4.75 at $c=1.0$. 
Fig.~\ref{fig:case_c} presents qualitative results across different $c$ values, demonstrating that smaller $c$ values (e.g., $c=0.0$) provide stronger protection but compromise visual quality, while larger $c$ values (e.g., $c=1.0$) lead to friendly visibility with reduced protection.
Based on the analysis, we selected $c=0.1$ as the optimal value, balancing acceptable protection effectiveness with maintainable visual quality.
}


\noindent \textbf{Sensitivity Analysis of Parameter $\alpha$.}
{Fig.~\ref{fig:hyp_alpha}~(a) illustrates the impact of $\alpha$ on visual quality and protection effectiveness. Increasing $\alpha$ from 0 to 0.005 sharply increases FID from 0 to ~33.93 (indicating more noticeable perturbations) while decreasing LIQE-SUM from ~7.56 to ~4.39 (indicating enhanced protection). Further increases to $\alpha=0.020$ continue improving FID to ~44.21 but result in saturated protection effectiveness (LIQE-SUM 3.55). Fig.~\ref{fig:hyp_alpha}~(b) shows that larger $\alpha$ values (e.g., 0.015, 0.020) require significantly more training iterations (1000) to converge compared to smaller values (e.g., $\alpha=0.005$ requiring ~80 iterations).
We set $\alpha=0.005$, which balances among visual quality, protection effectiveness, and computational efficiency.}

Note that Fig.~\ref{fig:hyp_alpha} uses $c=0.005$ to clearly demonstrate the relationship between $\alpha$ and training iterations, as $\alpha$ values at $c=0.1$ remain around 0.005 across the 240-1200 iteration range, making visualization challenging.
In practice, we adjust training steps to facilitate parallel acceleration to influence $\alpha$.
\begin{figure}
    \centering
    \includegraphics[width=\linewidth]{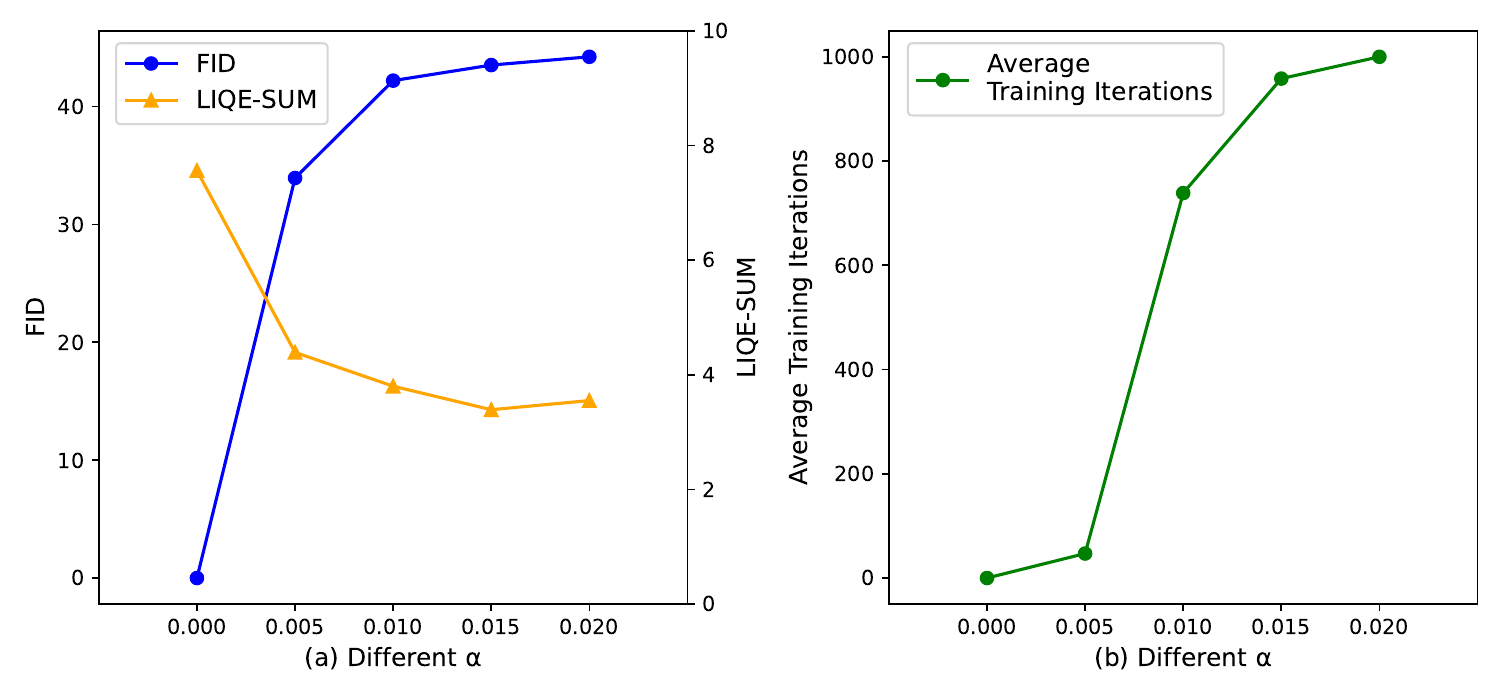}
    \vspace{-8mm}
    \caption{Sensitivity analysis of hyper-parameter $\alpha$. FID is calculated between adversarial data and corresponding clean data; here LIQE-SUM is the sum LIQE of Textual Inversion and DreamBooth, which are all the smaller the better. The dataset is VGGFace2 and $\epsilon=8/255$.}
    \label{fig:hyp_alpha}\vspace{-4mm}
\end{figure}

\section{Analysis of Adversary Settings}
Following Anti-DreamBooth~\cite{le_etal2023antidreambooth}, we consider three settings in this paper: ``convenient setting'', ``adverse setting'', and ``uncontrolled setting''.
\begin{figure*}
    \centering
    \includegraphics[width=\linewidth]{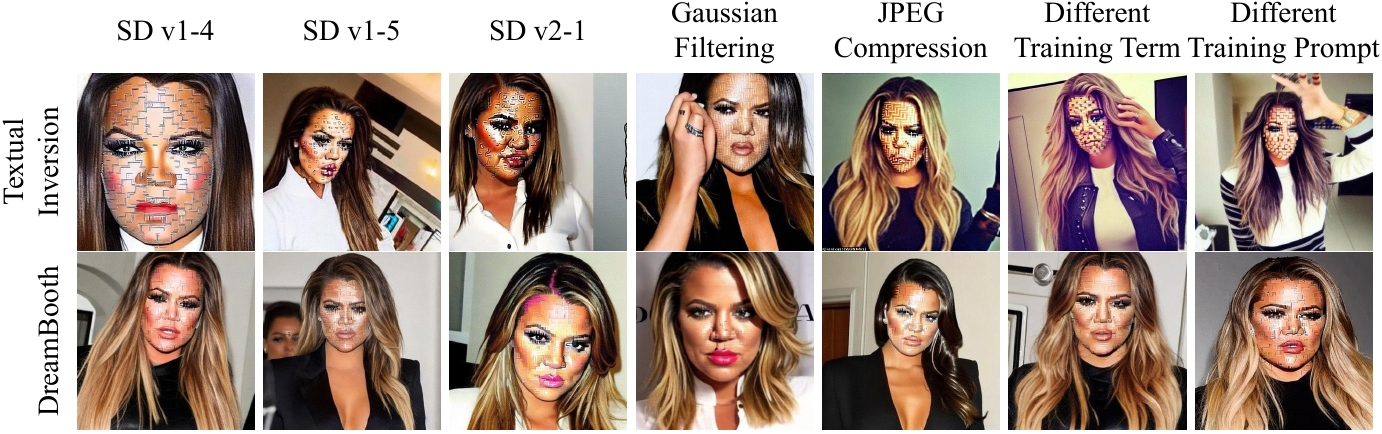}
    \caption{Experiments about different adversarial settings. Protector uses Stable Diffusion v1-4. Zoom in for a better view.}
    \label{fig:case_robustness}
\end{figure*}

\textbf{Convenient Setting.}
In a convenient setting (white-box setting), we (protectors) have all knowledge of the resources of adversary, including a pre-trained text-to-image generator, training term (e.g., ``sks''), and training prompt.
This scenario is practical because adversaries are likely to use high-quality, open-source pre-trained generators, with Stable Diffusion being the most prominent publicly available model.
Additionally, people often rely on open-source code with default training terms and prompts.	
The convenient setting is the simplest for the protectors, and all relevant experiment results have been analyzed and mentioned in our Experiment Section.

\begin{table}[]
    \centering 
     \caption{Results of model mismatching setting with the corresponding mean and standard deviation ($\pm$). ↑ denotes that higher values are preferable, while ↓ indicates the opposite. The best result in each column is highlighted in bold. SD is short for Stable Diffusion.}
    \label{ex:model_mismatching}
    \resizebox{0.45\textwidth}{!}{
    \begin{tabular}{ c c cc  cc}
    \toprule
    \multirow{2}{*}{\textbf{Protector}} &\multirow{2}{*}{\textbf{Adversary}}&
    \multicolumn{2}{c}{\textbf{Textual Inversion}}&
     \multicolumn{2}{c}{\textbf{DreamBooth}}\\
     ~&~&\textbf{LIQE~$\downarrow$} &\textbf{CLIP-FACE~$\downarrow$}&
     \textbf{LIQE~$\downarrow$} &\textbf{CLIP-FACE~$\downarrow$}\\
     \midrule
   No Defense&SD v1-4   &3.61\(_{\pm0.98}\) & 0.24\(_{\pm0.34}\)& 3.95\(_{\pm1.11}\) & 0.38\(_{\pm0.25}\) \\
   \midrule
   SD v1-4 &SD v1-4 & 2.31\(_{\pm0.78}\)&0.03\(_{\pm0.32}\) &2.06\(_{\pm0.70}\)&0.21\(_{\pm0.34}\)\\

   SD v1-4&SD v1-5& 2.06\(_{\pm0.74}\)&\textbf{0.02}\boldmath{\(_{\pm0.32}\)}&2.00\(_{\pm0.57}\)&0.24\(_{\pm0.34}\) \\
    SD v1-4&SD v2-1 base & \textbf{1.82}\boldmath{\(_{\pm0.60}\)}&0.06\(_{\pm0.30}\)&\textbf{1.70}\boldmath{\(_{\pm0.49}\)}&\textbf{0.20}\boldmath{\(_{\pm0.34}\)}\\
    \bottomrule
    \end{tabular}
    }\vspace{-4mm}
\end{table}

\begin{table*}[]
    \centering 
     \caption{Results of training term mismatching setting with the corresponding mean and standard deviation ($\pm$).
     ↑ means the higher the better, and vice versa. The best result in each column is in bold.}
    \label{ex:term_mismatching}
    \resizebox{0.9\textwidth}{!}{
    \begin{tabular}{c c cc  cc}
    \toprule
    \multirow{2}{*}{\textbf{Protector}} &\multirow{2}{*}{\textbf{Adversary}}&
    \multicolumn{2}{c}{\textbf{Textual Inversion}}&
     \multicolumn{2}{c}{\textbf{DreamBooth}}\\
     ~&~&\textbf{LIQE~$\downarrow$} &\textbf{CLIP-FACE~$\downarrow$}&
     \textbf{LIQE~$\downarrow$} &\textbf{CLIP-FACE~$\downarrow$}\\
     \midrule
  No Defense&sks&3.61\(_{\pm0.98}\) & 0.24\(_{\pm0.34}\)& 3.95\(_{\pm1.11}\) & 0.38\(_{\pm0.25}\) \\
   \midrule
    sks &sks & 2.31\(_{\pm0.78}\)&\textbf{0.03\(_{\pm0.32}\)} &2.06\(_{\pm0.70}\)&0.21\(_{\pm0.34}\)\\

   sks&t@t& \textbf{2.03}\boldmath{\(_{\pm0.74}\)} &0.04\(_{\pm0.34}\) &\textbf{1.93}\boldmath{\(_{\pm0.66}\)} &\textbf{0.21}\boldmath{\(_{\pm0.33}\)}\\
    \bottomrule
    \end{tabular}
    }\vspace{-4mm}
\end{table*}

\begin{table*}[]
    \centering 
     \caption{Results of training prompt mismatching setting with the corresponding mean and standard deviation ($\pm$).
     ↑ means the higher the better, and vice versa. The best result in each column is in bold.}
    \label{ex:prompt_mismatching}
    \resizebox{\textwidth}{!}{

    \begin{tabular}{ c c cc  cc}
    \toprule
    \multirow{2}{*}{\textbf{Protector}} &\multirow{2}{*}{\textbf{Adversary}}&
    \multicolumn{2}{c}{\textbf{Textual Inversion}}&
     \multicolumn{2}{c}{\textbf{DreamBooth}}\\
    ~&~&\textbf{LIQE~$\downarrow$} &\textbf{CLIP-FACE~$\downarrow$}&
     \textbf{LIQE~$\downarrow$} &\textbf{CLIP-FACE~$\downarrow$}\\
     \midrule
   No Defense&``a photo of sks person.''&3.61\(_{\pm0.98}\) & 0.24\(_{\pm0.34}\)& 3.95\(_{\pm1.11}\) & 0.38\(_{\pm0.25}\) \\
   \midrule
   ``a photo of sks person.''&``a photo of sks person.''& 2.31\(_{\pm0.78}\)&\textbf{0.03}\boldmath{\(_{\pm0.32}\)}&\textbf{2.06}\boldmath{\(_{\pm0.70}\)}&\textbf{0.21}\boldmath{\(_{\pm0.34}\)}\\

   ``a photo of sks person.''&``a dslr portrait of sks person.''& \textbf{2.23}\boldmath{\(_{\pm0.77}\)}&0.04\(_{\pm0.32}\)&2.62\(_{\pm1.18}\)&0.22\(_{\pm0.32}\)\\
    \bottomrule
    \end{tabular}
    }\vspace{-4mm}
\end{table*}

\begin{table*}[]
    \centering 
     \caption{Results of different post-processing with the corresponding mean and standard deviation ($\pm$).
     ↑ means the higher the better, and vice versa. The best result in each column is in bold.}
    \label{ex:post-processing}
    \resizebox{0.9\textwidth}{!}{

    \begin{tabular}{ c c cc  cc}
    \toprule
    \multirow{2}{*}{\textbf{Protector}} &\multirow{2}{*}{\textbf{Adversary}}&
    \multicolumn{2}{c}{\textbf{Textual Inversion}}&
     \multicolumn{2}{c}{\textbf{DreamBooth}}\\
    ~&~&\textbf{LIQE~$\downarrow$} &\textbf{CLIP-FACE~$\downarrow$}&
     \textbf{LIQE~$\downarrow$} &\textbf{CLIP-FACE~$\downarrow$}\\
     \midrule
   No Defense&Null&3.61\(_{\pm0.98}\) & 0.24\(_{\pm0.34}\)& 3.95\(_{\pm1.11}\) & 0.38\(_{\pm0.25}\) \\
   \midrule
   Null&Null& 2.31\(_{\pm0.78}\)&\textbf{0.03}\boldmath{\(_{\pm0.32}\)}&\textbf{2.06}\boldmath{\(_{\pm0.70}\)}&\textbf{0.21}\boldmath{\(_{\pm0.34}\)}\\
   Null&Gaussian Filtering &2.56\(_{\pm0.78}\) &0.18\(_{\pm0.31}\) &2.18\(_{\pm0.44}\)&0.31\(_{\pm0.34}\)\\
   Null&JPEG Compression &\textbf{2.08}\boldmath{\(_{\pm0.79}\)} &0.12\(_{\pm0.33}\) &2.09\(_{\pm0.46}\)&0.29\(_{\pm0.32}\)\\
    \bottomrule
    \end{tabular}
    }
    \vspace{-4mm}
\end{table*}

\begin{table*}[]
    \centering 
    \caption{Results of different uncontrolled settings with the corresponding mean and standard deviation ($\pm$).
    ↑ means the higher the better, and vice versa. The best result in each column is in bold.}
    \label{ex:uncondition}
    \resizebox{0.9\textwidth}{!}{

    \begin{tabular}{cc cc  cc}
    \toprule
    \multirow{2}{*}{\textbf{Perturbed}} &\multirow{2}{*}{\textbf{Clean}}&
    \multicolumn{2}{c}{\textbf{Textual Inversion}}&
     \multicolumn{2}{c}{\textbf{DreamBooth}}\\
     ~&~&\textbf{LIQE~$\downarrow$} &\textbf{CLIP-FACE~$\downarrow$}&
     \textbf{LIQE~$\downarrow$} &\textbf{CLIP-FACE~$\downarrow$}\\
     \midrule
   0&100\% &3.61$_{\pm0.98}$ & 0.24$_{\pm0.34}$& 3.95$_{\pm1.11}$ & 0.38$_{\pm0.25}$ \\
   \midrule
   25\% &75\%&2.82$_{\pm1.12}$ &0.20$_{\pm0.34}$ & 3.50$_{\pm1.25}$& 0.40$_{\pm0.28}$\\
   50\%&50\%& 2.15$_{\pm0.81}$&0.05$_{\pm0.35}$&2.89$_{\pm1.21}$&0.35$_{\pm0.32}$\\
    75\%&25\%&\textbf{2.03\boldmath{$_{\pm0.74}$}}& 0.04$_{\pm0.34}$&2.71$_{\pm1.07}$&0.33$_{\pm0.32}$\\
  \midrule
   100\% &0\% & 2.31$_{\pm0.78}$&\textbf{0.03\boldmath{$_{\pm0.32}$}} &\textbf{2.06\boldmath{$_{\pm0.70}$}}&\textbf{0.21}\boldmath{$_{\pm0.34}$}\\
    \bottomrule
    \end{tabular}
    }
    \vspace{-4mm}
\end{table*}

\textbf{Adverse Settings.}
In adverse settings, also known as grey-box settings, protectors lack knowledge of the specific version of the pre-trained text-to-image generator, training term, or training prompt employed by the adversary.
This scenario more closely resembles real-world conditions.
To defend against such attacks, protectors typically use surrogate modules to test the transferability of adversarial perturbations.
Experiments under both assumptions demonstrate the practicality of our method in real-world scenarios.
All experiments utilize $\epsilon$ = 8/255 as the perturbation size on the VGGFace2 dataset, with Stable Diffusion v1-4 as the base model, except in the case of model mismatching.

\begin{itemize}
    \item Model Mismatching. In this scenario, protectors use Stable Diffusion v1-4 without knowing the specific version of Stable Diffusion employed by the adversary. Results in Fig.~\ref{fig:case_robustness} and Table~\ref{ex:model_mismatching} show effective facial area protection even if adversaries used Stable Diffusion v1-4/v1-5/v2-1, which reflects the transferability of VCPro.
    \item Training Term Mismatching. We set protectors to use ``sks'' as a training term in Textual Inversion and DreamBooth, and adversaries use ``t@t''. As shown in Fig.~\ref{fig:case_robustness} and Table~\ref{ex:term_mismatching}, VCPro is robust to different training terms.
    \item Training Prompt Mismatching. We set protectors to use ``a photo of sks person.'' as training prompt in Textual Inversion and  DreamBooth, and adversaries to use ``a dslr portrait of sks person.''. As shown in Fig.~\ref{fig:case_robustness} and Table~\ref{ex:prompt_mismatching}, VCPro can still successfully protect the important concepts.
    \item Post-Processing. In this setting, we assume that the adversaries will use some post-processing to reduce the impact of the protected images, taking the two common post-processing techniques, Gaussian filtering and JPEG compression, as an example. Here we set the kernel size $7 \times 7$ for Gaussian filtering and adopt 75\% quality compression for JPEG, which also mentioned in~\cite{sandoval2023jpeg} can be a good purification method. As shown in Fig.~\ref{fig:case_robustness} and Table~\ref{ex:post-processing}, although the protective effect is somewhat diminished, the final generated image is still visibly cracked. VCPro can still successfully protect the target area.

\end{itemize}
\textbf{Uncontrolled Settings.}
In this section, we examine the scenario where an adversary has access to a set of original images of the target concept, which are then combined with protected images to train DreamBooth.
We evaluate three settings in which the proportion of original images is varied from 25\% to 75\%, as detailed in Table~\ref{ex:uncondition}.
All experiments employ a perturbation size of $\epsilon$ = 8/255 on the VGGFace2 dataset and utilize Stable Diffusion v1-4 as the base model.
Our protection remains effective as long as the proportion of protected images exceeds 50\%, but its effectiveness diminishes as the proportion of original images increases. Ideally, VCPro would receive platform support, become widely adopted, and be implemented across all social media platforms, thereby mitigating the risks associated with these uncontrolled settings.

\begin{figure*}[htbp]
\centering
\includegraphics[width=0.8\textwidth]{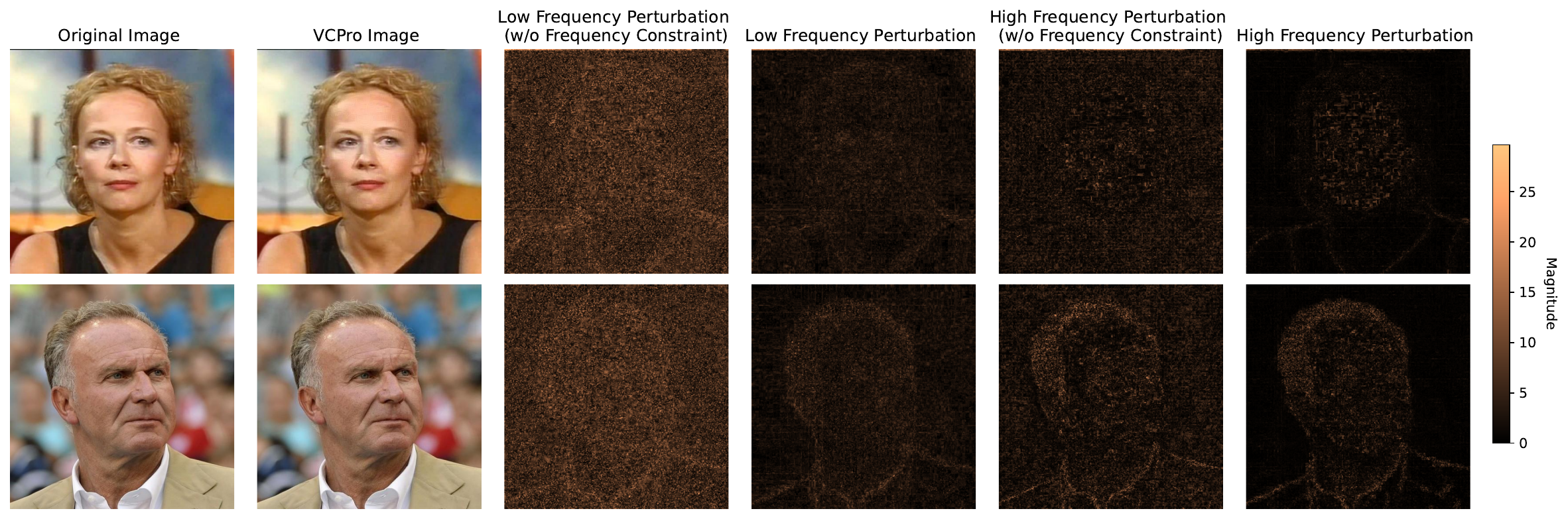}
\vspace{-4mm}
\caption{Frequency domain visualization of perturbations. From left to right: (1) Original image, (2) VCPro protected image, (3) Perturbation without our frequency constraint in low-frequency, (4) Perturbation with our frequency constraint in low-frequency, (5) Perturbation without our frequency constraint in high-frequency, (6) Perturbation with our frequency constraint in high-frequency. The color bar indicates the magnitude of perturbation.}
\label{fig:freq_visualization}
\end{figure*}
\begin{table}[htbp]
\centering
\caption{Results of frequency analysis. The dataset is VGGFace2, and the perturbation size is $8/255$. ↑ denotes that higher values are preferable, while ↓ indicates the opposite. Average Low/High Frequency Difference is calculated based on the average L1 norm per pixel. The best result in each column is highlighted in bold.}
\label{tab:freq_stats}
\begin{tabular}{@{}l@{\;}c@{\;}c@{}}
\hline
Metric & VCPro & ~~~w/o Freq. Constr. \\
\hline
Avg. Low Freq. Diff. ($\downarrow$) & \textbf{6.80} & 17.39 \\
Avg. High Freq. Diff. ($\downarrow$) & \textbf{6.98} & 11.47 \\
Low Freq. Ratio (\%) ($\downarrow$) & \textbf{49.37} & 60.26 \\
High Freq. Ratio (\%) ($\uparrow$) & \textbf{50.63} & 39.74 \\
\hline
\end{tabular}
\vspace{-6mm}
\end{table}

\section{Frequency Domain Analysis}
{We analyzed the frequency characteristics of VCPro via Discrete Wavelet Transform (DWT) with Haar wavelet decomposition, consistent with our perturbation generation process. This analysis is significant as human visual perception exhibits more sensitivity in low-frequency components than in high-frequency details.
We compared two variants: VCPro with frequency constraint, and without frequency constraint where $D(\cdot) = ||{x}-{x^\prime}||_{2}^{2}$ in Eq.~(8).
Further implementation details are provided in the supplementary materials.}

\noindent \textbf{Visualization in Different Frequency Components.}
{Fig.~\ref{fig:freq_visualization} shows the frequency visualization of sample images. The frequency-constrained variant shows significantly attenuated low-frequency perturbations (column 4) compared to the unconstrained variant (column 3); and concentrates perturbations predominantly in high-frequency components (column 6), with perturbation patterns that correspond to facial contours and edge features. Conversely, the unconstrained variant exhibits perturbation distribution across both frequency domains, with notably higher concentration in visually sensitive low-frequency regions (columns 3 and 5). Despite the protective modifications, the perceptual difference between original and protected images remains minimal.}

\noindent \textbf{Statistical Analysis in Different Frequency Components.} {Table~\ref{tab:freq_stats} provides a statistical analysis to quantify the frequency distribution of perturbations using the VGGFace2 dataset.
With our frequency constraint, both low-frequency (6.80) and high-frequency (6.98) differences are substantially lower than without the constraint (17.39 and 11.47, respectively).
Meanwhile, the frequency constraint effectively shifts perturbation distribution, reducing low-frequency components from 60.26\% (unconstrained) to 49.37\% (constrained).
Both findings effectively explain the improved visual quality of the constrained approach as low-frequency perturbations are more perceptible to human vision.}

\begin{figure*}
    \centering
    \includegraphics[width=\linewidth]{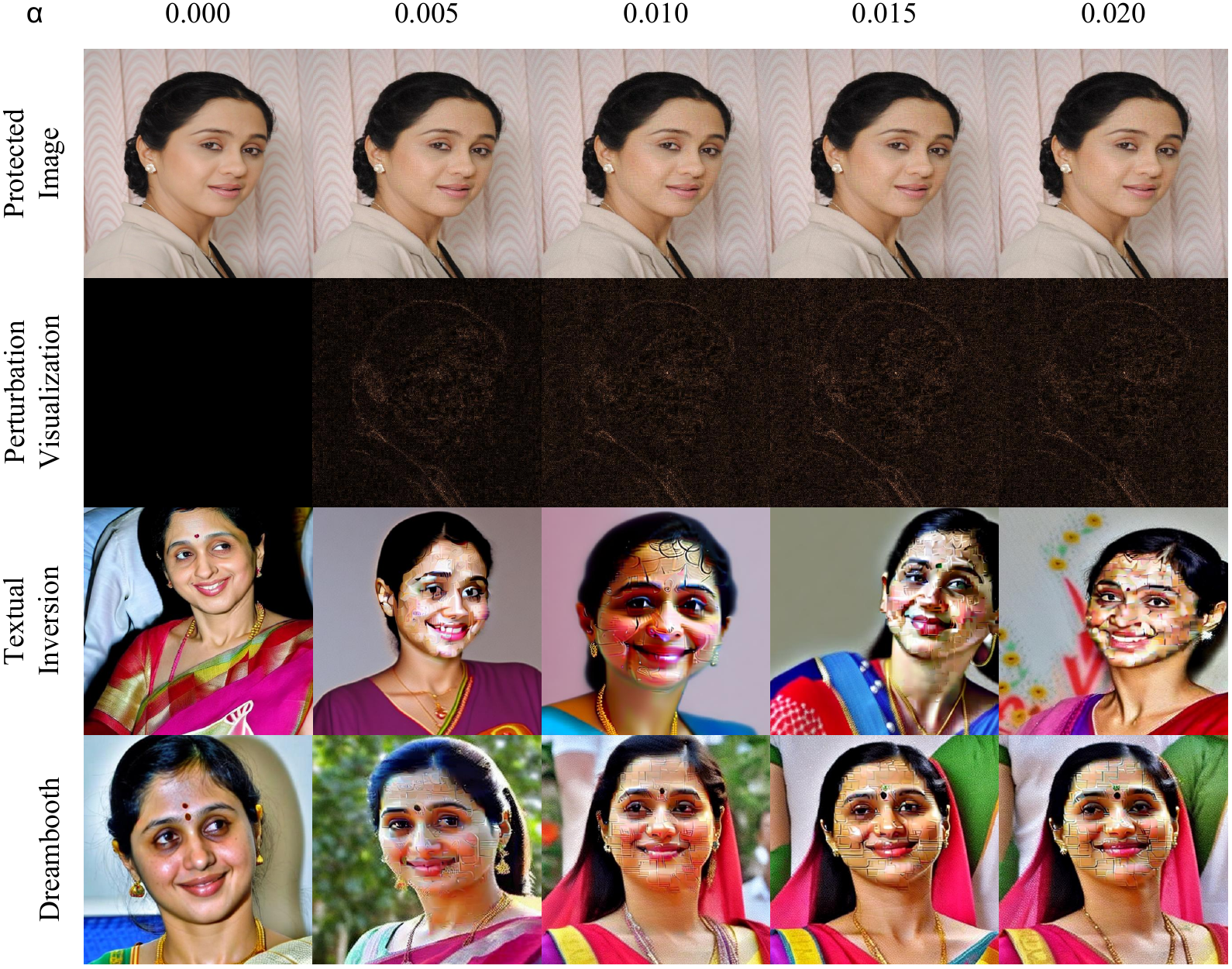}
    \caption{Qualitative results for different $\alpha$. For generating protected images, we use an $\epsilon$ value of 8/255 with Stable Diffusion v1-4. For better detail, please zoom in.}
    \label{fig:case_alpha}
    \vspace{-4mm}
\end{figure*}

\noindent \textbf{Sensitivity Analysis of Parameter $\alpha$.}
{Fig.~\ref{fig:hyp_alpha}~(a) illustrates the impact of $\alpha$ on visual quality and protection effectiveness. Increasing $\alpha$ from 0 to 0.005 sharply increases FID from 0 to ~33.93 (indicating more noticeable perturbations) while decreasing LIQE-SUM from ~7.56 to ~4.39 (indicating enhanced protection). Further increases to $\alpha=0.020$ continue improving FID to ~44.21 but result in saturated protection effectiveness (LIQE-SUM 3.55). Fig.~\ref{fig:hyp_alpha}~(b) shows that larger $\alpha$ values (e.g., 0.015, 0.020) require significantly more training iterations (1000) to converge compared to smaller values (e.g., $\alpha=0.005$ requiring ~80 iterations). Qualitative results in Fig.~\ref{fig:case_alpha} show similar trends. We set $\alpha=0.005$, which balances among visual quality, protection effectiveness, and computational efficiency.}

{Note that Fig.~\ref{fig:hyp_alpha} uses $c=0.005$ to clearly demonstrate the relationship between $\alpha$ and training iterations, as $\alpha$ values at $c=0.1$ remain around 0.005 across the 240-1200 iteration range, making visualization challenging.
In practice, to facilitate parallel acceleration, we adjust training steps to influence $\alpha$.
}
\begin{figure}
    \centering
    \includegraphics[width=\linewidth]{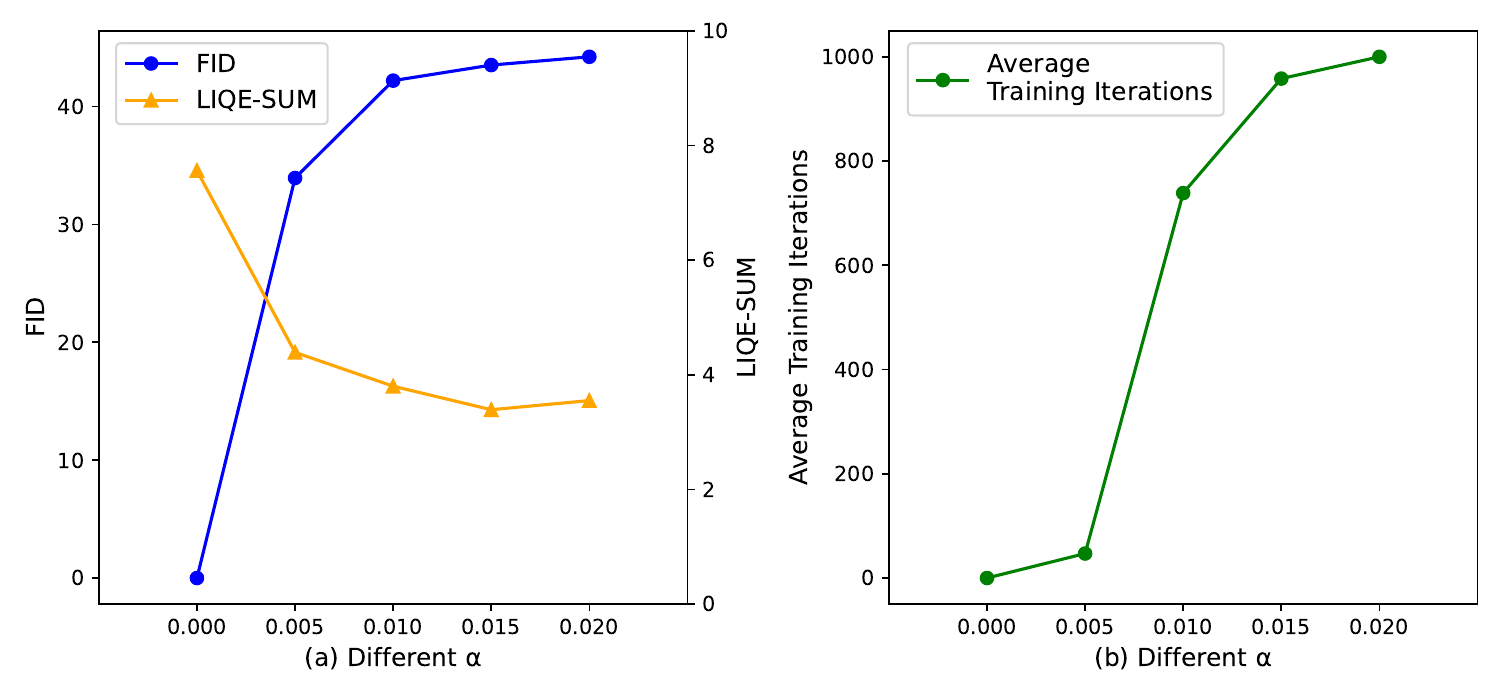}
    \vspace{-8mm}
    \caption{Sensitivity analysis of hyper-parameter $\alpha$. FID is calculated between adversarial data and corresponding clean data; here LIQE-SUM is the sum LIQE of Textual Inversion and DreamBooth, which are all the smaller the better. The dataset is VGGFace2 and $\epsilon=8/255$.}
    \label{fig:hyp_alpha}\vspace{-4mm}
\end{figure}
\section{Details of Frequency Analysis}
For each RGB image $I$, we performed the following analysis:

\begin{enumerate}
    \item Apply 2D DWT to each color channel $c \in \{R, G, B\}$ separately:
    $$\text{DWT}_c = \text{dwt2}(I_c, \text{`haar'})$$

    \item Extract the approximation coefficients (low-frequency component) $cA_c$ and detail coefficients (high-frequency components) $cH_c$, $cV_c$, and $cD_c$:
    $$\text{DWT}_c = (cA_c, (cH_c, cV_c, cD_c))$$

    \item Compute the combined high-frequency magnitude for each channel:
    $$H_c = \sqrt{cH_c^2 + cV_c^2 + cD_c^2}$$

    \item Calculate the frequency differences between clean image $I$ and protected image $I'$:
    \begin{itemize}
        \item Low-frequency difference: $\Delta L_c = |cA'_c - cA_c|$
        \item High-frequency difference: $\Delta H_c = |H'_c - H_c|$
    \end{itemize}

    \item Compute the mean differences across all channels:
    $$\Delta L = \frac{1}{3}\sum_{c \in \{R,G,B\}} \Delta L_c$$
    $$\Delta H = \frac{1}{3}\sum_{c \in \{R,G,B\}} \Delta H_c$$

    \item Calculate the ratio of perturbation in each frequency band:
    $$\text{Low Frequency Ratio} = \frac{\text{Mean}(\Delta L)}{\text{Mean}(\Delta L) + \text{Mean}(\Delta H)} \times 100\%$$
    $$\text{High Frequency Ratio} = \frac{\text{Mean}(\Delta H)}{\text{Mean}(\Delta L) + \text{Mean}(\Delta H)} \times 100\%$$
\end{enumerate}

The visualizations were then created using Matplotlib with a unified color scale (copper colormap) to ensure fair comparison between different noise components.

\begin{figure*}
    \centering
    \includegraphics[width=\linewidth]{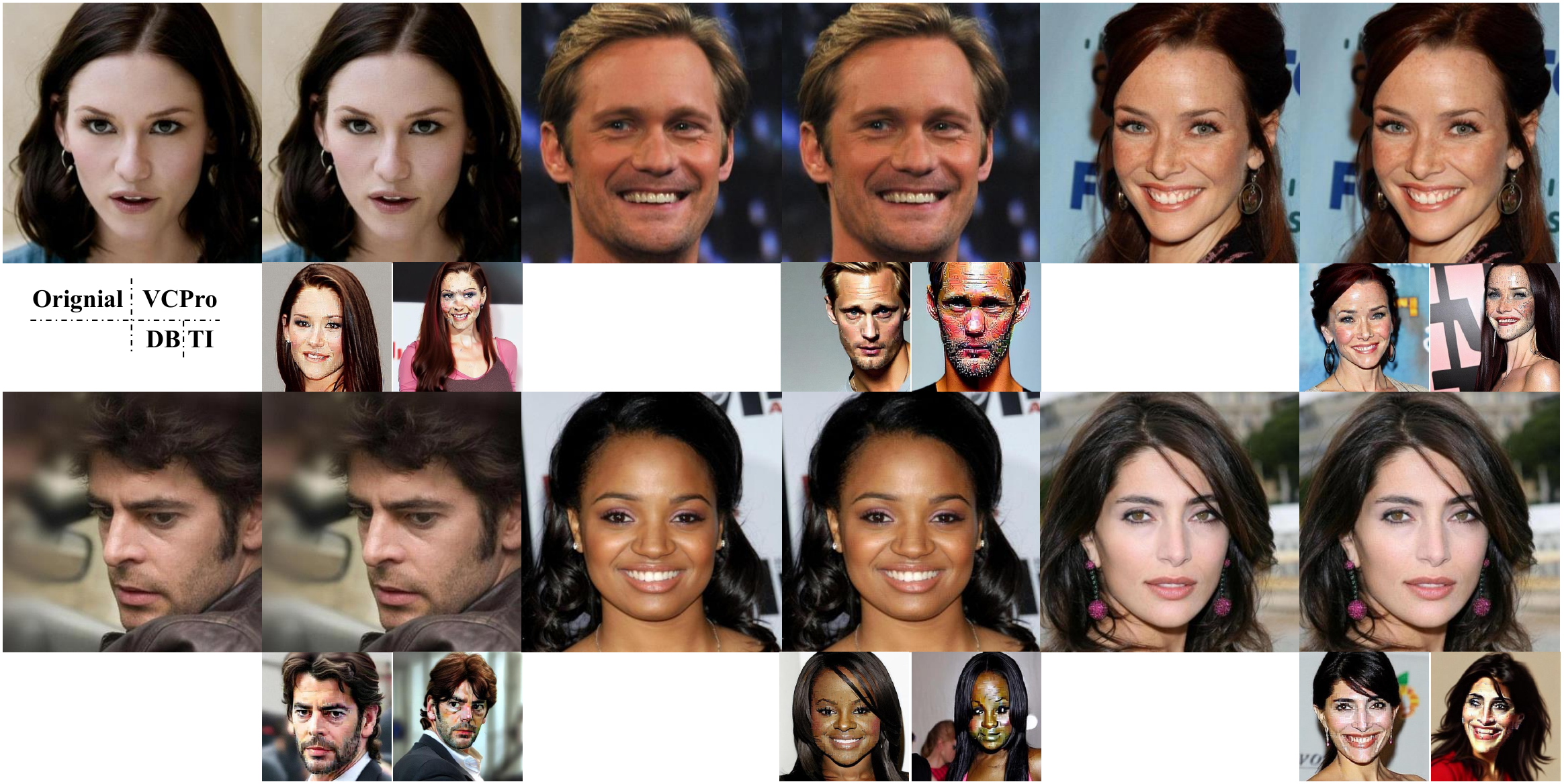}
    \vspace{-8mm}
    \caption{More qualitative defense results. For generating protected images, we use an $\epsilon$ value of 8/255 with Stable Diffusion v1-4. Each row displays the original images (No Defense), the protected images generated our method, along with respective outcomes in Textual Inversion~(TI) and DreamBooth~(DB). For better detail, please zoom in.}
    \label{fig:case_human}
    \vspace{-4mm}
\end{figure*}
\section{Case Show}
We show more cases of VCPro in Fig.\ref{fig:case_human}. VCPro can generate visually friendly perturbations while maintaining protection effectiveness.

\end{document}